\documentclass[a4paper]{article}
\usepackage{acl}

\usepackage{latexsym}
\usepackage{microtype}

\usepackage{amsmath}
\usepackage{amssymb}

\usepackage{booktabs}
\usepackage{multirow}
\usepackage{makecell}

\usepackage[ruled,vlined,linesnumbered]{algorithm2e}
\SetAlgoNlRelativeSize{0}

\usepackage{graphicx}
\usepackage{tikz}
\usepackage{pgfplots}
\pgfplotsset{compat=1.18}
\usetikzlibrary{positioning, arrows.meta, calc}

\usepackage{float}   
\usepackage{xspace}
\usepackage{url}
\usepackage{enumitem} 
\usepackage{pifont}  
\usepackage{alphalph} 
\newcommand{\cmark}{\ding{51}}
\newcommand{\xmark}{\ding{55}}

\title{Format-Constraint Coupling in Knowledge Graph Construction\\from Matrix-Layout Statistical Tables}

\author{
  Jingxuan Qi, Zhiqiang Ye, Yuxiang Feng\thanks{Corresponding author.} \\
  South China University of Technology \\
  Guangzhou, China \\
  \texttt{1312750677@qq.com, 1529890027@qq.com, yxfeng@scut.edu.cn}
}

\begin{document}
\maketitle

\begin{abstract}
An extraction schema should not reduce knowledge graph fidelity.
On statistical CSV, however, it can. We study
country$\times$year time-series matrices, a common layout on
open-data portals. In this setting, serialization format and schema
constraints interact \textit{super-additively}. Their joint effect
exceeds the sum of independent effects by up to $+$1.180
(2$\times$2 factorial, 6 datasets). Bootstrap 95\% CIs are strictly
positive on 4/6 datasets, with strongest evidence on wide Type-II
matrices. More critically, a schema applied to a mismatched format can
trigger \textbf{catastrophic mismatch}. Fact coverage falls below the
unconstrained baseline on 4/6 datasets through entity inflation or
extraction refusal. We call this observed pattern
\textbf{format-constraint coupling}. Probing and token ablation support
a surface-form anchoring explanation centred on column-name references.
Controlled variants across format-schema pairings, GraphRAG hosts,
and LLM families show the same direction within the measured scope;
one LLM family shows only partial activation. The observation also has
a diagnostic consequence. Three standard retrieval modes largely mask
construction quality ($\Delta{\leq}1$pp), whereas direct graph access exposes gaps up
to $+$47.6pp ($p{<}0.0001$). To support fidelity-aware evaluation, we
release CSVFidelity-Bench. It contains 15 datasets, 11 Type-II
matrices, 4 Type-III tables, and 1,892 Gold Standard facts across 6
domains.
\end{abstract}

\section{Introduction}
\label{sec:intro}

Applying an extraction schema to a knowledge graph construction
pipeline should not reduce fidelity. On statistical CSV, however, it
can. On World Bank population data, injecting a schema into
naive CSV chunks drops fact coverage from 18.7\% to 0.7\%.
Prior work \citep{Sui2024,Hegselmann2023} treats format and schema as
independent factors. Our experiments show a different pattern:
format and schema can interact \textit{super-additively}, with joint
effects exceeding the sum of individual contributions by up to $+$1.180.
We refer to this as \textbf{format-constraint coupling}.

GraphRAG systems \citep{Edge2024,Guo2024} assume unstructured text input.
Naive serialization of matrix-layout CSV fragments column headers,
breaking subject--time--value binding during graph construction.
Of 150 Gold Standard population facts, only 18.7\% survive intact.
This \textbf{cell-fact binding failure} occurs in the indexing layer,
distinct from table QA \citep{Herzig2020} or query-time reasoning
\citep{Chen2024tablerag,Zou2025}. Once binding breaks during construction,
no downstream path can recover it.

Deterministic parsers achieve FC$\geq$0.96 on 5/7 datasets at zero LLM cost.
The practical scope for LLM extraction is narrow. This paper asks when
LLM-based, schema-guided construction is helpful, when harmful, and how
to diagnose the difference. We use SGE (Structure-Guided Extraction) as
a controlled apparatus.
We make three contributions:

\begin{itemize}[nosep]
\item \textbf{Observed coupling.}
  A 2$\times$2 factorial ablation shows format and schema interact
  super-additively ($\Delta_\text{int}$ up to $+$1.180): mismatched
  schemas amplify errors (FC drops below baseline on 4/6 datasets).
  Token ablation supports surface-form anchoring.

\item \textbf{Boundary conditions.}
  Strongest on wide Type-II matrices, partial on Type-III, absent on
  long-format. Standard retrieval modes mask construction differences
  ($\Delta{\leq}1$pp), while direct graph access exposes gaps up to
  $+$47.6pp, showing E2E QA alone is insufficient.

\item \textbf{CSVFidelity-Bench.}
  15 datasets, 1,892 Gold Standard facts across 6 domains with
  deterministically verifiable ground truth.
\end{itemize}

\section{Related Work}
\label{sec:related}

\textbf{GraphRAG systems.} GraphRAG systems \citep{Edge2024,Guo2024,Gutierrez2024,Soman2024,Peng2024,Pan2024} add knowledge graphs to traditional RAG \citep{Lewis2020} but assume unstructured text input. Our experiments show MS GraphRAG and HippoRAG also fail on statistical CSV under default ingestion.

\textbf{Table-aware RAG.} TableRAG \citep{Chen2024tablerag}, RAG over Tables \citep{Zou2025}, and StructGPT \citep{Jiang2023structgpt} focus on query-time reasoning, not offline fact binding during graph construction.

\textbf{Schema-guided extraction.} REBEL \citep{CabotNavigli2021}, AutoSchemaKG \citep{Bai2025}, SPIRES \citep{Caufield2024}, GoLLIE \citep{Sainz2024gollie}, and CodeKGC \citep{Bi2024codekgc} target natural language text. AutoSchemaKG reaches FC=0.860--1.000 but inflates graphs (9K--12K nodes).

\textbf{Table understanding.} Table understanding methods \citep{Herzig2020,Liu2022,Yin2020,Zhang2023,Su2024,Wang2024cot,Hegselmann2023} address QA without persistent graphs. \citet{Sui2024} find Markdown outperforms CSV, but our Table~\ref{tab:cross-format} shows format advantages are \emph{contingent on schema alignment}: Markdown yields FC=1.000 under matched schema but FC=0.000 under mismatch. Column-level methods \citep{Hulsebos2019,Zhang2020web,Deng2022,Dong2023} and Table-to-KG pipelines \citep{Dimou2014,Shraga2020,JimenezRuiz2020} address typing or RDF mapping. W3C CSVW requires pre-authored metadata, unsuitable for zero-shot ingestion.

\textbf{Constrained generation.} Constrained decoding \citep{Willard2023,Zheng2023} and structured prompting \citep{Khattab2023,BeurerKellner2023,Liu2024instructor} ensure syntactic compliance. Our finding is orthogonal: even with perfect syntax, failure can be \textit{semantic binding}.

\textbf{Distinction from format sensitivity.} \citet{Sclar2024} and \citet{Lu2022} document quantitative sensitivity. Our finding is qualitatively different: format and schema interact \textit{multiplicatively}, producing emergent failures (entity proliferation, refusal) that neither factor predicts alone.

\section{Experimental Apparatus}
\label{sec:method}

The pipeline is a lightweight experimental apparatus for manipulating
format and schema independently. The coupling pattern it reveals
(\S\ref{sec:coupling}) is the contribution, not the pipeline itself.

\textbf{Definitions.} A statistical CSV file $T$ contains deterministic
facts $\mathcal{F} = \{(s_i, t_i, v_i)\}$ (subject, time, value).
A graph construction system turns $T$ into $G = (V, E)$.
\textbf{Fact coverage}
$\text{FC} = |\{f \in \mathcal{F} \mid f \sqsubseteq_2 G\}| /
|\mathcal{F}|$ measures the proportion of Gold Standard triples
correctly bound within 2-hop neighbourhoods; \textbf{entity coverage}
(EC) measures Gold Standard entities found in the graph.

\begin{figure*}[htb]
  \centering
  \includegraphics[width=\textwidth]{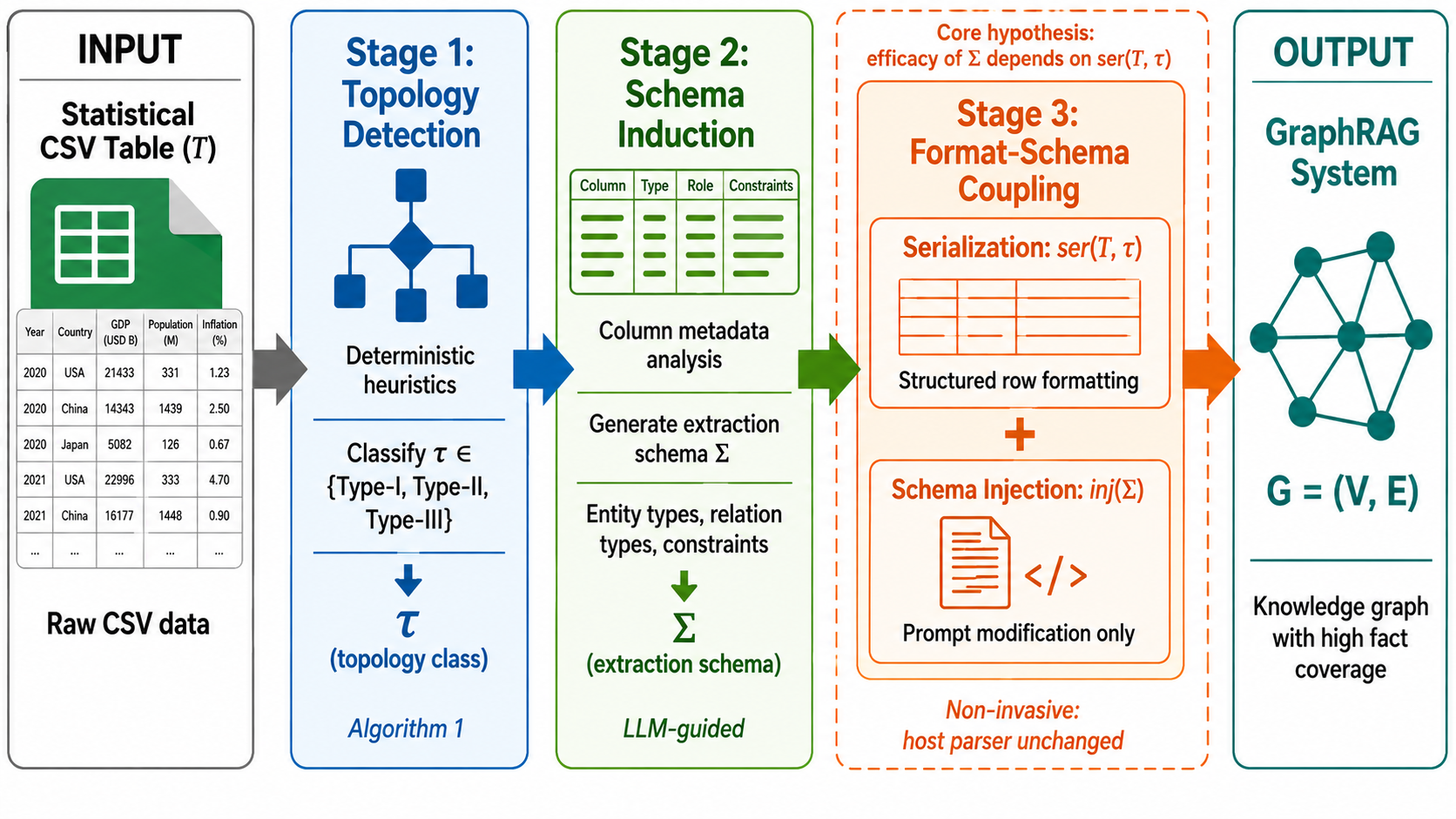}
  \caption{SGE three-stage apparatus for statistical CSV.}
  \label{fig:pipeline}
\end{figure*}

\textbf{Three-stage apparatus}
(Figure~\ref{fig:pipeline}). We insert a structure-aware apparatus
upstream of the GraphRAG index:
$T \xrightarrow{\text{S1}} (\tau, \mathcal{S})
\xrightarrow{\text{S2}} \Sigma \xrightarrow{\text{S3}}
(\text{ser}(T,\tau),\, \text{inj}(\Sigma)) \!\rightarrow\! G$.
\textit{Stage~1} classifies the CSV topology
$\tau \in \{\text{Type-I},\, \text{Type-II},\, \text{Type-III}\}$
via deterministic heuristics (Appendix~\ref{app:alg1}).
\textit{Stage~2} induces an extraction schema
$\Sigma$ (entity types, relation types, and extraction constraints)
from column metadata (Appendix~\ref{app:c}).
\textit{Stage~3} serializes input into structured rows
($\mathrm{ser}(T,\tau)$) and injects the schema into the host
system's prompt ($\mathrm{inj}(\Sigma)$)---only the prompt changes;
host parsers and storage remain untouched.
\textbf{Core hypothesis}: the efficacy of $\Sigma$ depends on whether
$\mathrm{ser}(T,\tau)$ provides anchorable row-level structure
(\S\ref{sec:coupling}).

\textbf{Adaptive degradation.}
\label{sec:adaptive}
Two safeguards preserve worst-case parity with baseline.
First, small Type-III datasets ($n_\text{rows}{<}20$) skip schema
injection. Second, if post-extraction edge/node ratio drops below
$\theta{=}0.90$, the system reverts to baseline mode. This proxy
cannot detect value-level failures: graphs may appear structurally
normal while values are incorrectly bound. We implement on
LightRAG \citep{Guo2024}; cross-host validation on MS GraphRAG confirms
portability (Appendix~\ref{app:q}).

\section{Experiments}
\label{sec:experiments}

We organize experiments as follows. \S\ref{sec:protocol} defines the
evaluation protocol. \S\ref{sec:main-results} reports main results and
baselines. \S\ref{sec:coupling} presents factorial evidence for
format-constraint coupling. \S\ref{sec:mismatch} analyses the
behavioural signatures of mismatch and anchoring.
\S\ref{sec:downstream} tests downstream observability.
\S\ref{sec:robustness} examines robustness.

\subsection{Evaluation Protocol}
\label{sec:protocol}

\textbf{Evaluation metrics.} We use two primary metrics.
\textbf{Entity coverage (EC)} is the substring-match proportion of
Gold Standard entities found in the graph. \textbf{Fact coverage (FC)}
is the proportion of Gold Standard triples captured within 2-hop
neighbourhoods. FC naturally favours schema-normalised entity names. A
de-biased evaluation confirms that the naming-bias net effect is
$\leq$1.6\% (\S\ref{sec:robustness}). As a complementary metric,
\textbf{canonical triple F1} normalises Gold Standard and system outputs
to $(s_\text{norm}, t, v_\text{norm})$ triples. It then computes
Precision, Recall and F1. This measure is independent of graph topology
and comparable to standard IE tasks.

\textbf{CSVFidelity-Bench.} We build a diagnostic benchmark targeting
cell-fact binding failure: 15 datasets from 4 sources across 6
domains, with 1,892 Gold Standard facts (11 Type-II, 4 Type-III;
Table~\ref{tab:main}). Of these, 10 were used during development;
5 are held-out. Deterministic scripts generate the majority of gold
facts (96.1\% inter-annotator agreement on manual subsets). The
benchmark measures \textbf{transport fidelity}: the proportion of
CSV facts correctly bound in the graph. Full construction
methodology, topology distribution, and composition analysis are
in Appendix~\ref{app:k}.

\textbf{Experimental configuration.} We compare SGE against LightRAG
Baseline using Claude Haiku 4.5 at temperature 0 and mxbai-embed-large
embeddings. Cross-system comparisons cover LightRAG v1.4.12,
MS GraphRAG, HippoRAG v2, nano-GraphRAG, AutoSchemaKG and
RAG-Anything. Across systems, three independent metrics show consistent
effect directions: FC, canonical triple F1 and graph traversal accuracy
(Appendix~\ref{app:b}). Permutation tests are the primary inference
method because facts from the same entity share an LLM reasoning context.
This violates independence. Wilcoxon, Bootstrap and McNemar tests serve
as robustness checks (Appendix~\ref{app:b}). Table~\ref{tab:main}
evaluates on 50 countries. Table~\ref{tab:ablation} uses a 25-country
subset. Minor Baseline FC differences between the tables reflect
sampling, not methodological inconsistency (Appendix~\ref{app:50c}).

\subsection{Main Results}
\label{sec:main-results}

\begin{table}[htb]
\centering
\caption{Cross-dataset FC results (2-hop). Top block: standard
comparisons; middle block: OOD expansion; bottom block: long-format
boundary cases. $\dagger$Adaptive degradation (\S\ref{sec:adaptive}).
$\S$50-country evaluation; Table~\ref{tab:ablation} uses the
25-country subset (a strict subset of these 50), so Baseline FC may
differ slightly between the two tables. *Manual override
(Appendix~\ref{app:the}).}
\label{tab:main}
\resizebox{\columnwidth}{!}{%
\begin{tabular}{llccrrr}
\toprule
\textbf{Dataset} & \textbf{Dom.} & \textbf{Type} & \textbf{Gold}
  & \textbf{SGE FC} & \textbf{Base FC} & \textbf{Ratio} \\
\midrule
Annual Budget       & Gov  & II    & 4e/20f   & \textbf{1.000} & 1.000 & 1.00$\times$ \\
Food Safety$^\dagger$ & Gov & III  & 17e/52f  & \multicolumn{2}{c}{\textit{degraded}} & N/A \\
Health Statistics   & Gov  & II-T  & 3e/14f   & \textbf{0.786} & 0.643 & 1.22$\times$ \\
Inpatient Stat.     & Gov  & III   & 8e/16f   & \textbf{0.938} & 0.438 & \textbf{2.14$\times$} \\
WHO Life Expect.$^\S$    & Gov  & II    & 50e/300f & \textbf{1.000} & 0.170 & \textbf{5.88$\times$} \\
WB Child Mortality$^\S$  & Gov  & II    & 50e/300f & \textbf{1.000} & 0.433 & \textbf{2.31$\times$} \\
WB Population$^\S$       & Gov  & II    & 50e/300f & \textbf{1.000} & 0.133 & \textbf{7.52$\times$} \\
WB Maternal Mort.$^\S$   & Gov  & II    & 50e/300f & \textbf{0.973} & 0.820 & \textbf{1.19$\times$} \\
Fortune 500 Rev.    & Fin  & II    & 25e/125f & \textbf{1.000} & 0.400 & \textbf{2.50$\times$} \\
THE Univ.\ Ranking  & Acad & III$\to$II* & 25e/150f & \textbf{0.600} & 0.207 & \textbf{2.90$\times$} \\[-2pt]
\multicolumn{7}{l}{\footnotesize *Algorithm~1 classifies THE as Type-III ($|C_\text{key}|{=}2$, $|C_T|{=}0$); manual override to Type-II because} \\[-3pt]
\multicolumn{7}{l}{\footnotesize \phantom{*}scores are flat-ranked, not hierarchically nested (Appendix~\ref{app:g}).} \\
\midrule
\multicolumn{7}{l}{\textit{Out-of-domain expansion (unseen during design)}} \\
WB Cereal Prod.     & Agri & II    & 25e/40f  & \textbf{0.950} & 0.050 & \textbf{19.0$\times$} \\
WB CO$_2$ Emissions & Env  & II    & 25e/40f  & \textbf{0.700} & 0.025 & \textbf{28.0$\times$} \\
WB Pop.\ Growth     & Demo & II    & 25e/40f  & \textbf{0.625} & 0.075 & \textbf{8.33$\times$} \\[1pt]
\multicolumn{7}{l}{\footnotesize\textit{OOD datasets: 40 facts each, unseen during design; same config as main experiments.}} \\
\midrule
\multicolumn{7}{l}{\textit{Long-format Type-III (scope boundary)}} \\
Eurostat Crime      & Gov  & III-L & 35e/105f & 0.000          & \textbf{0.410} & 0.00$\times$ \\
US Census Demo.     & Gov  & III-L & 30e/90f  & \textbf{0.244} & 0.022 & \textbf{11.0$\times$} \\
\bottomrule
\end{tabular}%
}
\end{table}

In the 13 standard comparisons in Table~\ref{tab:main}, SGE FC is at
least as high as Baseline in 12 cases. It is strictly higher in 11.
This count excludes Food Safety, which triggers adaptive degradation,
and Eurostat Crime, which is analysed as a long-format boundary
failure. On the 50-country international datasets, Wilcoxon signed-rank
tests give $p{<}0.011$ for all 4 datasets (Appendix~\ref{app:b}). Effect
sizes are large: WHO $r{=}0.963$, WB~CM $r{=}0.797$, WB~Pop $r{=}0.963$,
WB~Mat $r{=}0.963$. The two long-format Type-III datasets mark a scope
boundary. Eurostat Crime has SGE FC=0.000, whereas Baseline reaches 0.410.
Long-format rows lack sufficient per-row density for coupling to activate
(Limitations~\ref{sec:limitations}).

\textbf{Concrete example.} On WHO Life Expectancy, Baseline extracts
28/150 facts correctly (FC=0.187). Common errors include: (1) binding
``China 2020'' to the wrong value (e.g., 2019's value); (2) creating
separate entities for each year (``2020'', ``2021'') instead of treating
them as temporal attributes; (3) dropping country--value associations
entirely when column headers are fragmented across chunks. Full SGE
reaches FC=1.000 by preserving row-level structure: each serialized row
reads ``Country\_Code: CHN | 2020: 77.4 | 2021: 78.2'', allowing the
schema to anchor ``Country\_Code'' as the subject and year columns as
temporal attributes.

\textbf{Cross-system comparison.} Five alternative GraphRAG systems
(MS~GraphRAG, LightRAG v1.4.12, HippoRAG~v2, nano-GraphRAG,
RAG-Anything) show limitations under default statistical-CSV ingestion,
with WHO FC ranging from 0.007 to 1.000
(Table~\ref{tab:cross-system} in Appendix~\ref{app:cross-system-table}).
This is a diagnostic default-ingestion comparison, not a tuned SOTA
benchmark. AutoSchemaKG reaches FC=0.860--1.000 on 4 datasets but at the
cost of severe graph inflation (9K--12K nodes vs 1.5K--4.5K for SGE).
MS~GraphRAG handles regular Type-II input (WHO FC=1.000) but fails on
Type-III (Inpatient FC=0.000), suggesting that default community detection
does not reliably recover hierarchical semantics.

\textbf{Supplementary baselines and scope}
(Appendix~\ref{app:m}).
Six alternative approaches (deterministic parser, row-local, fixed
schema, JSON~SO, few-shot, table-aware prompt) all show high
cross-dataset variance. Row-local (headers without schema) performs
\textit{worse} than Baseline on 4/7 datasets---headers alone cannot
substitute for coupling. A deterministic parser reaches
FC$\geq$0.96 on 5/7 well-structured datasets at zero LLM cost,
narrowing the practical scope for LLM extraction.  The coupling
finding stands independently: the factorial ablation
(\S\ref{sec:ablation}) isolates format-schema interaction in the
evaluated matrix-layout settings (canonical triple F1: SGE Recall=0.915 vs
Baseline 0.008; Appendix~\ref{app:b}).

\subsection{Factorial Evidence for Format-Constraint Coupling}
\label{sec:coupling}
\label{sec:ablation}

The main evidence for format-constraint coupling comes from a
2$\times$2 factorial test, followed by two supporting analyses that
help explain where the effect appears and how it behaves:
(1)~\textit{main effect}---does a super-additive interaction exist?
(2$\times$2 factorial, this section);
(2)~\textit{moderator identification}---which datasets are most
vulnerable? (column-name descriptiveness, this section);
(3)~\textit{targeted manipulation}---what kind of format--schema
matching best explains the observed failures? (probing $+$ token
ablation, \S\ref{sec:mismatch}).

\begin{figure*}[htb]
\centering
\includegraphics[width=\textwidth]{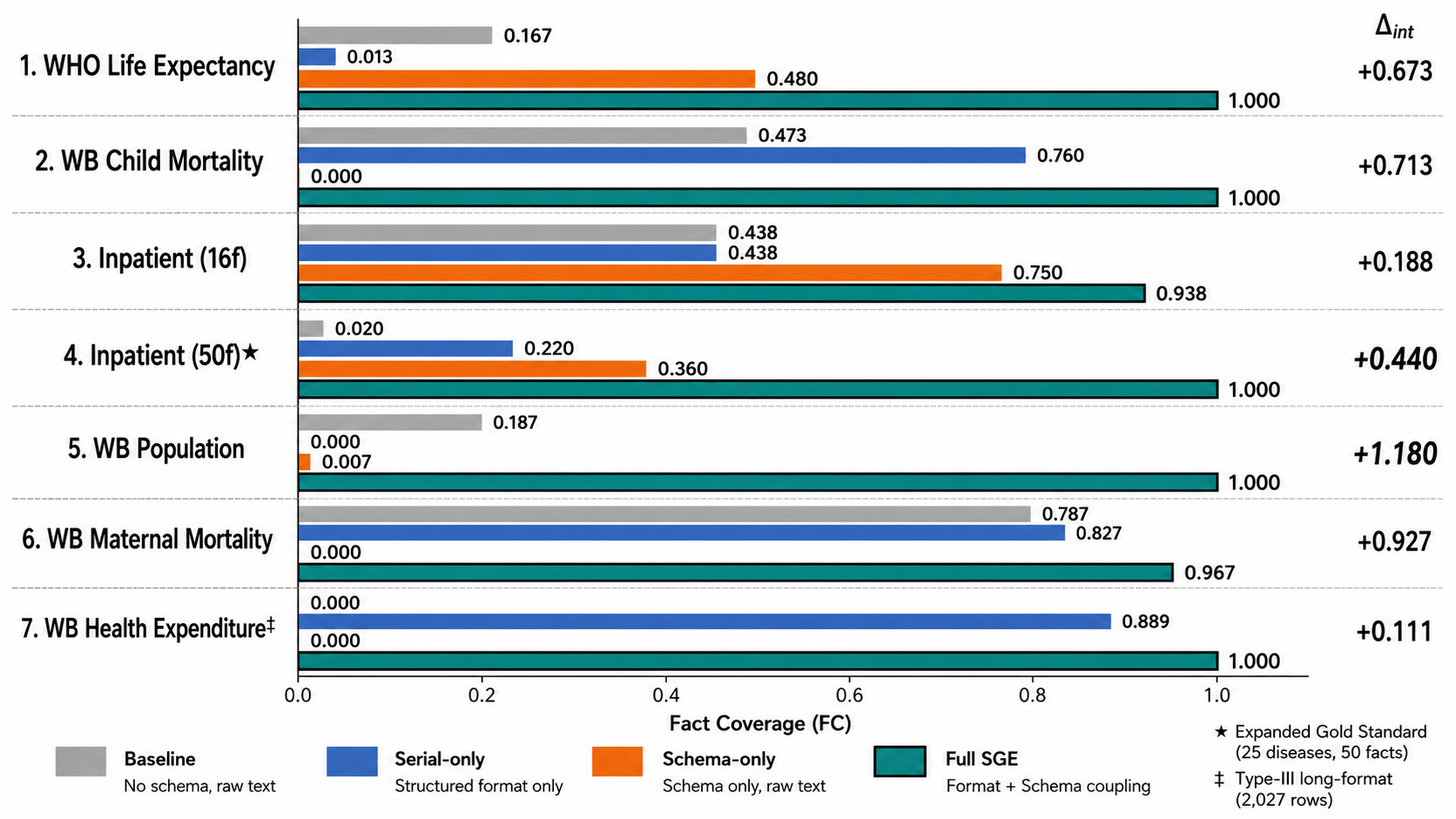}
\caption{2$\times$2 Factorial Ablation across 7 datasets (25-country subset).
$\star$Expanded Gold Standard (25 diseases, 50 facts).
$\ddagger$Type-III long-format (2,027 rows).}
\label{tab:ablation}
\end{figure*}

We use a 2$\times$2 factorial design on 6 datasets to separate
structured serialization from schema constraints. The four conditions
are \textit{Baseline}, \textit{Serial-only}, \textit{Schema-only} and
\textit{Full~SGE}. Baseline uses naive text chunking and the default
prompt. Serial-only uses SGE-serialized rows with the default prompt.
Schema-only uses naive text chunking with the SGE schema injected into
the entity extraction prompt. This replacement keeps the output-format
instructions and tuple delimiters unchanged. Only entity type
definitions and extraction rules differ from the default prompt. Full
SGE combines SGE-serialized rows with the same injected schema. Chunking
boundaries are determined by LightRAG. They are identical for conditions
that share the same input format. If format and schema were additive,
Schema-only should never fall \textit{below} Baseline. Table~\ref{tab:ablation}
shows that it does so on 4/6 datasets.

Let $\mathcal{F}$ and $\mathcal{S}$ denote the sets of available
serialization formats and extraction schemas, with $f_0 \in \mathcal{F}$
(naive chunking) and $s_0 \in \mathcal{S}$ (no schema) as baselines.
We define \textit{format-constraint coupling} as the phenomenon where
a coverage function $\text{FC}(f, s)$ exhibits a positive interaction term
$\Delta_\text{int} > 0$:
\begin{equation}\label{eq:coupling}
\Delta_\text{int} = \text{FC}(f,s) - \text{FC}(f,s_0) - \text{FC}(f_0,s) + \text{FC}(f_0,s_0)
\end{equation}
accompanied by qualitatively different LLM failure modes under
mismatch ($\text{FC}(f_0, s) < \text{FC}(f_0, s_0)$) rather than
gradual degradation. This is distinct from format sensitivity
\citep{Sclar2024}, which describes \textit{variance} across formats
($\text{Var}_f[\text{FC}(f, s)]$ for fixed~$s$); coupling is
a \textit{cross-derivative}
($\partial^2 \text{FC}/\partial f \, \partial s \neq 0$)
that produces behavioural phase transitions.

The interaction term (Eq.~\ref{eq:coupling}) is positive across all
6 datasets (Table~\ref{tab:ablation}). It ranges from $+$0.111
(WB Health Exp) to $+$1.180 (WB Pop). Values above 1 are possible when
both independent conditions fall \textit{below} Baseline. The joint
condition then recovers from a doubly negative state. This pattern
indicates a qualitative phase transition rather than gradual
improvement. For example, on WB Pop: Baseline FC=0.187, Serial-only
FC=0.000, Schema-only FC=0.007. Both individual interventions are
\textit{worse} than doing nothing. Yet Full SGE (combining both)
reaches FC=1.000, yielding $\Delta_\text{int}{=}+1.180$. This is not
additive improvement---it is a phase transition from catastrophic
failure to perfect recovery.

Bootstrap 95\% CIs are strictly positive on 4/6 datasets.
The Fisher combined test gives $p{<}0.001$. The remaining two datasets,
Inpatient and WB Health Exp, cross zero because they have small sample
sizes ($n{=}9$--$16$ facts). Per-dataset CIs are in
Appendix~\ref{app:50c}.

Extended 50-country validation uses 300 facts per dataset and no new
LLM runs. All four CIs are strictly positive (WHO [0.813, 0.853],
WB~CM [0.547, 0.587], WB~Pop [0.813, 0.853], WB~Mat [0.140, 0.180]).
The Fisher combined test again gives $p{<}0.001$ (Appendix~\ref{app:50c}).

The interaction terms reveal two key properties.

\textbf{Property 1: Schema effects depend strongly on format.} Schema-only
causes FC to collapse \textit{at or below} Baseline on 4/6 datasets
(Table~\ref{tab:ablation}). The remaining 2/6 (WHO FC=0.480,
Inpatient FC=0.360) have descriptive column names that provide
partial anchoring even in unstructured chunks. For example, WHO's
\texttt{Life\_expectancy} column is self-documenting, whereas WB Pop's
\texttt{SP.POP.TOTL} is opaque without external metadata.

A single moderating variable helps explain this split:
\textbf{column-name descriptiveness}, quantified as \textit{Column
Descriptiveness Score} (CDS; Eq.~\ref{eq:cds} in
Appendix~\ref{app:cds}), which captures the fraction and informativeness
of entity-identifying columns. A binary split at CDS$\geq$0.02 correctly
separates all 7 main datasets for entity-level mismatch: high-CDS
datasets (WHO, Inpatient) retain partial grounding under Schema-only,
while low-CDS datasets (all WB, CDS$\leq$0.008) collapse to
FC$\leq$0.007. This threshold emerges from the data rather than being
pre-specified.\footnote{The CDS--FC correlation $r{=}0.92$ is based on
$n{=}5$ factorial datasets and should be treated as descriptive rather
than predictive; see Appendix~\ref{app:cds} for expanded validation.}
Cross-dataset manipulation is consistent: the \textit{same} schema
perturbation (AY-3d) yields FC=1.000 on WHO but FC=0.000 on WB~CM
(Appendix~\ref{app:grounding-distance}).

Serial-only shows inconsistent effects. It improves WB~CM (FC=0.760 vs
Baseline 0.473) and WB~Mat (FC=0.827 vs Baseline 0.787), where structured
rows preserve in-row entity--value co-occurrence. It hurts WHO (FC=0.013
vs Baseline 0.167) and WB~Pop (FC=0.000 vs Baseline 0.187). A chunk-size
ablation rules out chunk capacity as the cause: doubling chunk size from
600 to 1200 tokens does not improve Serial-only performance on WHO
(Appendix~\ref{app:chunk-size}). The failure stems from absent schema
constraints, not insufficient context. Without schema guidance, the LLM
cannot determine which columns represent entities versus values.
Serialization alone is unreliable without a matched schema.

In sum, neither factor alone is reliably positive. Schema-only
collapses on 4/6 datasets. Serial-only improves only 2/6. Their
combination succeeds on all 6. WB Pop exemplifies the crossover. Both
individual conditions fall \textit{below} Baseline (Serial-only FC=0.000,
Schema-only FC=0.007), yet the joint condition reaches FC=1.000
($\Delta_\text{int}{=}+1.180$; Figure~\ref{fig:crossover} in
Appendix~\ref{app:50c}).

\textbf{Intuition.} In naive chunking, a chunk reads
``\textit{CHN 1411778724 71.4 ...}''---numbers without column labels.
Structured serialization adds labels
(``\textit{Country\_Code: CHN, 2020: 1411778724}''), but the generic
prompt does not know \texttt{Country\_Code} is a subject. Only when
\textit{both} format and schema are present can the LLM bind values:
the schema provides the \textit{what}; the format provides the
\textit{where}.

\textbf{Property 2: Mismatch triggers qualitative behavioural change,
not gradual decline.} In the Schema-only condition, FC falls below the
unconstrained Baseline on 4/6 datasets. This is not reduced precision.
It is a shift in LLM behaviour. WB Pop Schema-only produces 3.47$\times$
as many entities as Full SGE, reflecting ungrounded proliferation. WHO
shows 43.6\% empty responses, reflecting extraction refusal.
\S\ref{sec:mismatch} gives quantitative analysis of both failure modes.

\textbf{Scope and alternatives.} Coupling is robust for Type-II, where
4/4 CIs are strictly positive. Type-III is confirmed on Inpatient
($\Delta_\text{int}{=}+0.440$ [0.280, 0.600]). It is not confirmed on very
large tables where format's independent effect dominates
(Appendix~\ref{app:50c}). Cross-host replication, supplementary
baselines and Stage-2 ablation address three alternatives: prompt
conflict, chunking artefacts and format sufficiency
(Appendix~\ref{app:q}/\ref{app:m}/\ref{app:c}).

\subsection{Behavioral Analysis: Mismatch and Anchoring}
\label{sec:mismatch}

The Schema-only degradation in Table~\ref{tab:ablation} is not reduced
precision. It is a qualitative shift in LLM behaviour. When schema
$\Sigma$ is applied to naive CSV chunks, the LLM either promotes numeric
cells into separate entities or refuses to extract. We call this pattern
\textbf{catastrophic mismatch}. Well-formed constraints can amplify,
rather than suppress, extraction errors. The LLM nominally complies with
the schema, with 68.9\% type match. Yet without structured row-level
input, it cannot reliably ground entities to source rows. The entities
are correctly typed but semantically unbound. The contrast with Full SGE
(FC=1.000 under the identical schema) is consistent with mismatch
between schema and input format. It is not explained by contradictions
within the prompt itself.

\begin{figure*}[htb]
  \centering
  \begin{minipage}[t]{0.48\textwidth}
    \centering
    \includegraphics[width=\textwidth]{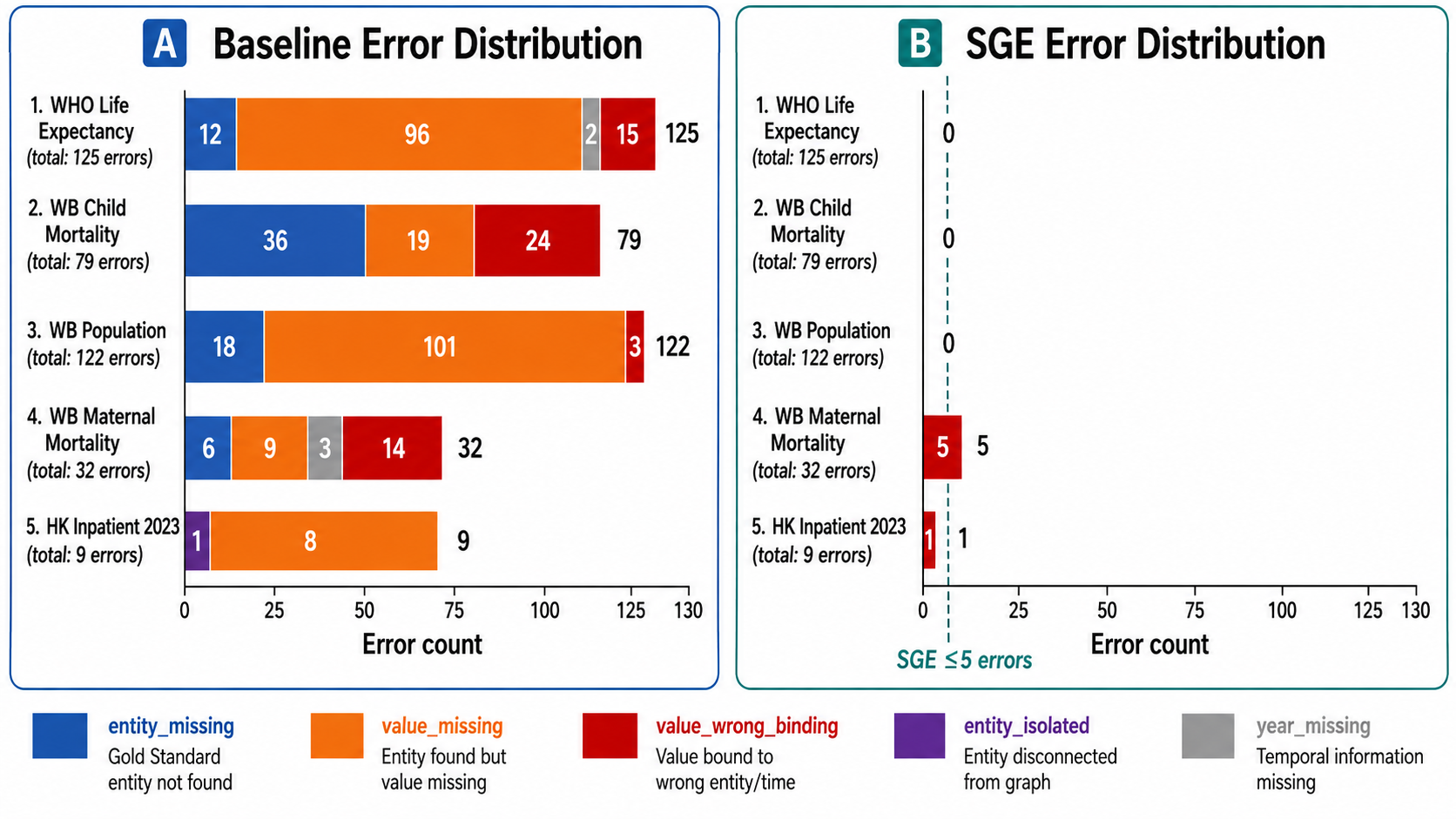}
    \caption{Error taxonomy across 5 datasets.}
    \label{fig:errors}
  \end{minipage}
  \hfill
  \begin{minipage}[t]{0.48\textwidth}
    \centering
    \includegraphics[width=\textwidth]{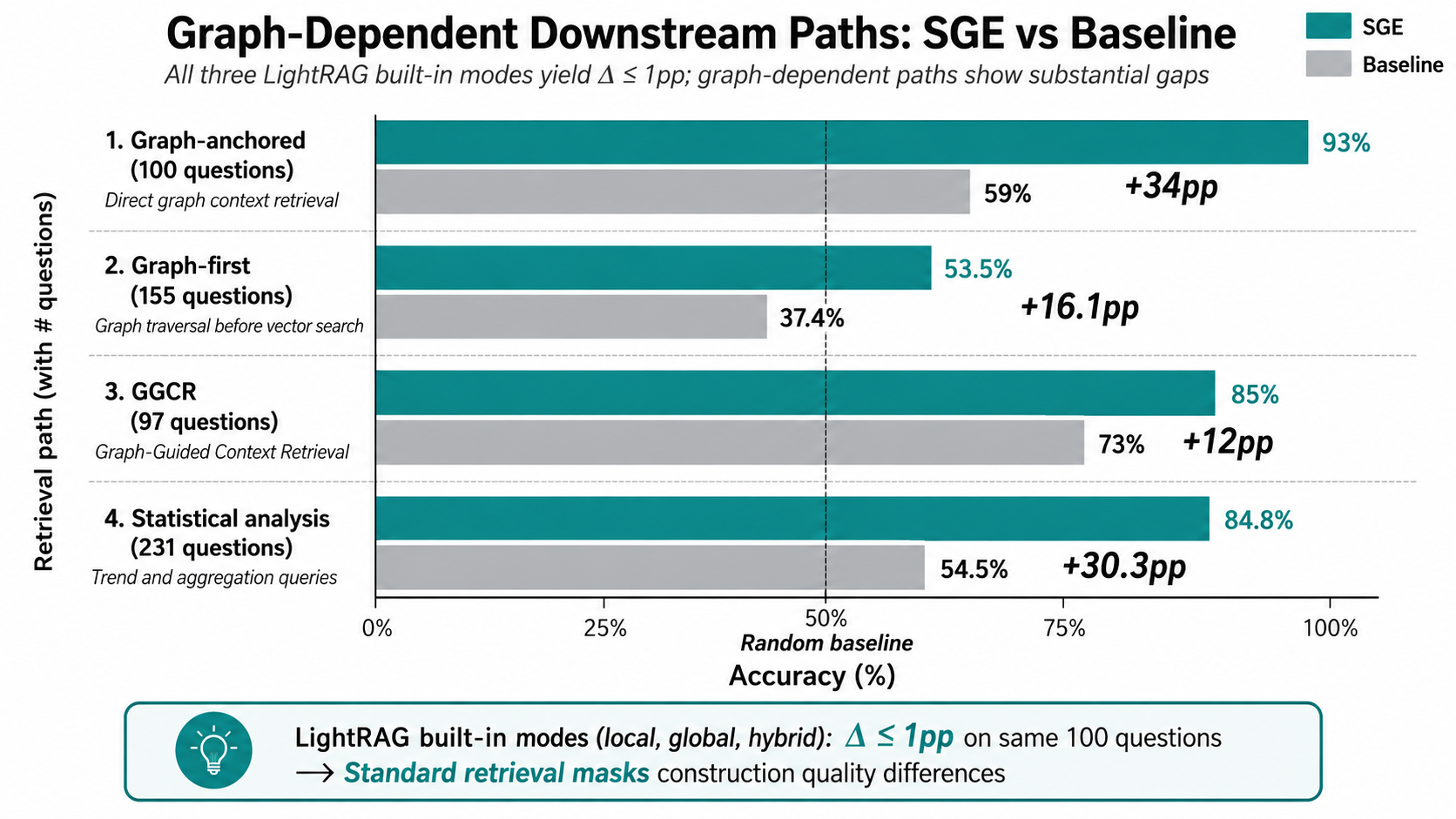}
    \caption{Graph-dependent downstream evaluation.}
    \label{tab:downstream}
  \end{minipage}
\end{figure*}

\textbf{Quantitative analysis.}
Two failure modes emerge.
\textit{Mode 1: Ungrounded entity proliferation} appears on WB Pop,
WB~CM and WB~Mat. Schema-only generates 15,782 entities, compared with
4,548 for Full SGE (3.47$\times$ increase). This occurs because the LLM
promotes numeric cells (e.g., population values like ``1411778724'') into
standalone entity nodes, unbound to any country despite 68.9\% nominal
schema compliance. The schema specifies entity types correctly, but
without structured row-level input, the model cannot determine which
numbers are values versus which are entity identifiers.
\textit{Mode 2: Extraction refusal} appears on WHO,
where 43.6\% of Schema-only responses are empty. The model abstains
rather than producing incorrect output; Full SGE has 0\% refusal.
All 6/6 datasets show one of these patterns (Appendix~\ref{app:p}).
Figure~\ref{fig:errors} visualises the error distribution.

\textbf{Cross-format validation.} One possible objection is that schema
failure on raw text is simply expected because the schema was designed
for SGE serialization. To probe that possibility, we run two controls:
(1) apply SGE schema to other formats (JSON records, Markdown tables)
and check for the same mismatch pattern; (2) include JSON Structured
Output (\S\ref{sec:main-results}, independently designed JSON Schema $+$
matching JSON format, WHO FC=1.000) as a positive instance of another
matched format-schema pairing.

\begin{table}[htb]
\centering
\caption{Cross-Format Validation: Format $\times$ Schema Matrix
(WHO, WB Pop, Inpatient). Matched pairings activate coupling;
mismatched pairings do not. Interact.\ column shows $+$Schema $-$
($-$Schema) for same-format rows (Appendix~\ref{app:s}).}
\label{tab:cross-format}
\footnotesize
\begin{tabular}{@{}llcrrc@{}}
\toprule
\textbf{Format} & \textbf{Schema} & \textbf{DS}
  & \makecell{\textbf{+Sch}\\\textbf{FC}}
  & \makecell{\textbf{$-$Sch}\\\textbf{FC}}
  & \textbf{$\Delta$} \\
\midrule
SGE ser.            & SGE             & WHO & \textbf{1.000} & 0.013 & $+$\textbf{.947} \\
JSON SO             & matched         & WHO & \textbf{1.000} & ---   & ---               \\
MD table            & matched         & WHO & \textbf{1.000} & 0.127 & $+$\textbf{.833} \\
Raw text            & SGE             & WHO & 0.207          & 0.167 & .000             \\
MD table            & SGE$^\dag$      & WHO & 0.000          & 0.127 & $-$\textbf{.167} \\
\midrule
SGE ser.            & SGE             & WBP & \textbf{1.000} & 0.000 & $+$\textbf{1.18} \\
MD table            & matched         & WBP & \textbf{0.920} & 0.000 & $+$\textbf{.920} \\
MD table            & matched         & Inp & \textbf{1.000} & 0.063 & $+$\textbf{.937} \\
\bottomrule
\multicolumn{6}{@{}l@{}}{\scriptsize $^\dag$SGE schema references ``Country\_Code'' (absent from MD);} \\
\multicolumn{6}{@{}l@{}}{\scriptsize column-name mismatch is the operational coupling mechanism.} \\
\end{tabular}
\end{table}

The critical difference between matched and mismatched schemas is the
column-name reference. The Markdown schema references ``Country Name'',
which is present in Markdown headers. The SGE schema references
``Country Code'', which is absent from Markdown output. This contrast
isolates column-name anchoring as the operational coupling locus.

Two conclusions follow from this supporting control. First, the same
mismatch pattern appears across three independent format-schema
pairings: SGE, JSON~SO and Markdown. It also appears on both Type-II and
Type-III examples. This argues against a serialization-specific
artefact. Second, the key difference is schema alignment. The same
Markdown format yields FC=1.000 under a matched schema and FC=0.000
under the SGE schema (Table~\ref{tab:cross-format}). The result depends
on whether column references remain alignable.

\textbf{Anchoring probing.} Table~\ref{tab:cross-format} shows that
format-schema matching is necessary, but does not reveal \textbf{where the
matching operates}. An 11-condition probing experiment
(Appendix~\ref{app:probing}, Table~\ref{tab:probing}) manipulates
individual schema components while holding SGE serialization fixed.
Table~\ref{tab:anchoring} summarizes the key conditions that reveal a
three-level anchoring hierarchy.

\begin{table}[htb]
\centering
\small
\caption{Anchoring hierarchy probing (WHO, 150 Gold Standard facts;
  SGE serialization fixed). Only active misdirection to a
  matchable-but-wrong column triggers catastrophic degradation.
  Full 11-condition table in Appendix~\ref{app:probing}.}
\label{tab:anchoring}
\begin{tabular}{llc}
\toprule
\textbf{Cond.} & \textbf{Intervention} & \textbf{FC} \\
\midrule
AX    & Correct column ref.\ (baseline)       & \textbf{1.000} \\
AY-2  & Semantic misdirection                  & \textbf{1.000} \\
AY-3d & Column ref.\ removed entirely          & \textbf{1.000} \\
AY-3  & Wrong-but-existing column ref.         & \textbf{0.400} \\
\bottomrule
\end{tabular}
\end{table}

Three findings emerge. First, \textbf{column-name references} are the
primary anchor. Redirecting a column reference to a \textit{wrong but
existing} column is the only intervention that triggers catastrophic
degradation. AY-3 reduces FC from 1.000 to 0.400 and replicates on 4
datasets. Second, entity type names provide auxiliary anchoring. Third,
semantic descriptions have negligible effect (AY-2). A targeted
grounding-distance experiment sharpens this result
(Appendix~\ref{app:grounding-distance}). Removing column references,
using positional descriptions or referencing a non-existent column all
preserve FC=1.000 on WHO. Only \textbf{active misdirection} to a
matchable-but-wrong anchor triggers collapse. On WB~CM, all non-exact
references yield FC=0.000. This pattern is consistent with
column-name descriptiveness moderating fallback grounding capacity.

\textbf{Behavioral signature: surface-form anchoring.} We
hypothesize that structured serialization creates repeating token
patterns that serve as anchors: the LLM matches schema field names
to column-label \textit{tokens} rather than semantic content.
This account is consistent with ICL findings
\citep{Min2022,Wei2023}; however, coupling differs in its behavioural
form: instead of smooth performance variation, mismatch produces
phase-transition-like failures. Our probing shows misdirected
\textit{semantic} descriptions have no effect (AY-2), while misdirected
\textit{structural} column references cause catastrophic failure
(AY-3). We quantify this via \textit{Template Token Fraction}
(TTF): TTF$\geq$0.4 holds for 85\% of successful matched extractions
(Appendix~\ref{app:ttf}).\footnote{AY-3d rules out internal schema
inconsistency: removing column references entirely preserves FC=1.000---only
active misdirection to a wrong-but-existing column triggers collapse
(Appendix~\ref{app:grounding-distance}).}
Summarizing the evidence chain:
\begin{enumerate}[nosep,leftmargin=1.5em]
\item Format-schema mismatch causes catastrophic FC drops
  (factorial, Table~\ref{tab:ablation}).
\item Column-name descriptiveness perfectly predicts which datasets
  are vulnerable (binary split, 7/7 correct).
\item Experimentally manipulating column references confirms the
  behavioural pattern (AY-3 series, Table~\ref{tab:anchoring}), and
  the \textit{same} manipulation yields opposite outcomes on
  descriptive vs.\ non-descriptive datasets
  (Appendix~\ref{app:grounding-distance}).
\end{enumerate}
\noindent Together, these three levels support column-name
anchoring as a primary locus of format-constraint coupling.

\textbf{Input-side token ablation.}
The probing experiments above manipulate \textit{schema} components. A
complementary experiment manipulates \textit{input} tokens while holding
the schema fixed. It uses 7 conditions (M0--M6) and 20 gold-filtered chunks per
dataset (Appendix~\ref{app:token-ablation}). Key findings:
(1) Masking column-label tokens (M1) has no effect on WHO (descriptive
columns), where FC is unchanged. The same intervention reduces WB~Pop FC
by 17\%, replicating CDS moderation from the schema side. This asymmetry
confirms that descriptive column names provide fallback grounding even
when labels are masked.
(2) Masking entity-name tokens (M3) collapses WHO FC by 95\%, from
0.400 to 0.020. The extraction structure remains intact, with 20
entities and 440 relations. Entity names are therefore necessary for
fact \textit{binding}, but not for coupling \textit{activation}.
(3) Crucially, shuffling field order within rows (M5) and row order within
chunks (M6) preserve FC on both datasets. These results rule out
positional pattern matching. They confirm that coupling operates via
\textbf{surface-form anchoring}: the LLM matches schema field names to
column-label \textit{tokens}, not recurring \textit{positions}. The model
relies on lexical overlap between schema field names and input tokens,
rather than learning positional regularities across rows.

\textbf{Cross-host validation.} Porting SGE to MS GraphRAG v3.0.8
(prompt template adaptation only) yields FC=1.000 on all 3 test datasets,
including Inpatient where GraphRAG Baseline FC=0.000
(Appendix~\ref{app:q}).

\subsection{Downstream Tasks: Retrieval Path Determines Fidelity Visibility}
\label{sec:downstream}

Coupling makes construction quality difficult to assess through
standard downstream evaluation alone. A schema can yield FC=1.000 under
a matched format and degrade to FC=0.000 under mismatch
(\S\ref{sec:ablation}). Retrieval may still appear stable. We therefore
ask which retrieval architectures make upstream fidelity differences
visible downstream. We use a diagnostic probing set of binding-sensitive
lookups and trend queries, not an open-ended QA benchmark.

All graph-dependent paths show substantial SGE advantages
(Table~\ref{tab:downstream}); gaps concentrate on binding-sensitive
queries (point lookup $+$47.6pp, trend $+$20.8pp;
Appendix~\ref{app:graph-first}).

\textbf{Naming bias control.}
Substring matching could favour SGE's schema-normalised names.
Three controls rule this out (Appendix~\ref{app:naming-bias}):
fuzzy matching \textit{widens} the gap ($+$34pp$\to$$+$35pp);
in 26/29 SGE-only correct cases, the Baseline also retrieved the
same entity but lacked values; and a value-first de-biased FC
shows naming bias $\leq$5.3\% across all 5 evaluated datasets,
with the net effect favouring the Baseline on 3/5.

\textbf{Standard retrieval masks fidelity differences.}
All three LightRAG built-in modes (naive, local, global) yield
$\Delta{\leq}1$pp on the same 100 questions. They are effectively blind
to construction quality when embedding-based entry points dominate. For
example, on WHO point lookups, naive mode achieves 59\% accuracy on both
SGE and Baseline graphs despite a 5.88$\times$ FC gap. This occurs because
vector retrieval can locate relevant chunks even when the graph contains
incorrect bindings. The retrieved text chunks still contain the correct
answer, masking the graph construction failure.

Graph-first retrieval is only \textbf{partially sensitive}. Its overall
gain is $+$16.1pp, concentrated on binding-dependent lookups ($+$47.6pp)
and trends ($+$20.8pp). Crucially, even graph-first retrieval shows near-zero
sensitivity on some query types: definition queries ($\Delta{=}0$pp) and
aggregation queries ($\Delta{=}+2$pp) are insensitive because they rely
on entity existence rather than precise value binding.

All 6 major GraphRAG systems we surveyed (LightRAG, MS~GraphRAG, HippoRAG,
nano-GraphRAG, RAPTOR, GraphReader) evaluate exclusively via E2E QA
(Appendix~\ref{app:eval-blindness-survey}). This makes construction-layer
regressions easy to miss in current practice. We demonstrate the
blindness directly on LightRAG's three modes. The survey suggests, but
does not prove, that similar evaluation habits persist elsewhere. The
implication is that construction fidelity requires direct measurement
(e.g., Gold Standard fact coverage) rather than relying on downstream
QA accuracy as a proxy.

\subsection{Robustness}
\label{sec:robustness}

\textbf{OOD stress test.} On 10 out-of-domain Type-II datasets spanning
8 new domains (agriculture, environment, demographics, education,
non-government sectors), SGE outperforms Baseline on 7/10. The median
improvement is $11.0\times$ (range: 8.33$\times$ to 28.0$\times$). For
example, WB CO$_2$ Emissions achieves SGE FC=0.700 vs Baseline FC=0.025
(28$\times$ improvement). The remaining 3 failures (Fortune 500 Revenue,
THE University Ranking, OECD Germany Health) are caught by the automatic
edge/node-ratio fallback, which correctly reverts to baseline mode when
e/n drops below 0.90. An OECD factorial provides additional evidence
for Schema-only catastrophic mismatch beyond World Bank data: OECD Germany
Schema-only FC=0.000 vs Full SGE FC=1.000
(Appendix~\ref{app:r},~\ref{app:oecd-factorial}).

\textbf{Cross-model validation.} Claude Haiku and GPT-5-mini both show
coupling on all 5 evaluated datasets. Gemini activates coupling on 3/5
but fails on 2/5 (WB~CM FC=0.020, WB~Mat FC=0.040). The failure reflects
insufficient \textit{compliance depth}: entities are generated
(EC=0.64--0.92), but relations are not correctly bound. Gemini uses
numeric literals as node IDs, causing cross-entity merging during
deduplication. Named Entity Connectivity Rate is 0.18 on failures and
$\geq$0.98 on successes (Appendix~\ref{app:cross-model}). Gemini
Schema-only controls confirm this: WHO Schema-only FC=0.613 exceeds
Full~SGE FC=0.493, showing structured format \textit{hinders} rather
than helps. Claude shows normal coupling: Schema-only FC=0.480 is below
Full~SGE FC=1.000. Coupling therefore does not reduce to instruction
following. GPT-5-mini Schema-only confirms catastrophic mismatch on
non-descriptive columns (Appendix~\ref{app:cross-model}).

\textbf{Blind test.} On 6 unseen datasets: SGE wins 2/6, ties 0/6,
Baseline wins 1/6, both FC=0.0 on 3/6. Dual failures share a common
profile: pure numeric columns without textual descriptors---consistent
with column-name descriptiveness as a pre-condition for entity-level
coupling (\S\ref{sec:ablation}; Appendix~\ref{app:g}).

\section{Conclusion}
\label{sec:conclusion}
\label{sec:discussion}

This paper reports two findings on statistical CSV ingestion for GraphRAG
systems. First, \textbf{format-constraint coupling} occurs on matrix-layout
statistical CSV tables. Serialization format and schema constraints can
interact super-additively: their joint effect exceeds the sum of individual
contributions by up to $+$1.180 (Eq.~\ref{eq:coupling}). Mismatch can
trigger qualitative behavioural shifts, including entity inflation
(3.47$\times$ node count increase) and extraction refusal (43.6\% empty
responses). These shifts differ from gradual degradation. The current
evidence is strongest on Type-II matrices (4/4 factorial CIs strictly
positive), partial on hierarchical Type-III tables (1/2 confirmed), and
negative on the two long-format cases tested.

Probing and token ablation support a \textit{surface-form} anchoring
account. The LLM matches schema field names to column-label \textit{tokens},
rather than to recurring positions or semantic descriptions. This
interpretation is consistent with ICL findings \citep{Min2022,Wei2023}.
It differs from format sensitivity \citep{Sclar2024,Lu2022}, which produces
quantitative variation rather than phase-transition-like failures.

Supporting replications across three format-schema pairings, two GraphRAG
hosts and two LLM families suggest that the pattern is not tied to a single
serialization or host design. A third LLM family shows only partial
activation.

Second, \textbf{evaluation blindness} follows from this construction
failure. All three standard retrieval modes are effectively blind to
construction quality ($\Delta{\leq}1$pp). Direct graph access exposes
gaps up to $+$47.6pp ($p{<}0.0001$). In the systems we surveyed,
evaluation is consistently framed around E2E QA. Such evaluation can
miss construction-layer regressions.

Four implications follow:
(1)~\textit{Route by quality}---deterministic parsing suffices for
well-structured tables (FC$\geq$0.96 on 5/7 datasets at zero LLM cost);
LLM extraction adds value mainly on wide matrices with descriptive columns
(CDS$\geq$0.02) and sufficient template token overlap (TTF$\geq$0.4;
Appendix~\ref{app:ttf}). For example, WHO Life Expectancy (26 columns,
CDS=0.020, TTF=0.42) benefits from LLM extraction (FC=1.000), whereas
Fortune 500 Revenue (5 columns, purely numeric) is better served by
deterministic parsing (FC=1.000 at zero cost).

(2)~\textit{Co-design format and schema}---mismatched schemas are
worse than none. On WB Pop, Schema-only FC=0.007 is 27$\times$ worse than
Baseline FC=0.187. This means value-level probes are needed after
construction to detect binding failures that graph-structure metrics miss.

(3)~\textit{Graph-structure metrics are insufficient}---edge/node ratio
detects catastrophic extraction collapse but misses value-level binding
failures. Eurostat Crime has normal graph topology (e/n=1.23) but zero
correct facts (FC=0.000).

(4)~\textit{Evaluate fidelity directly}---E2E QA misses binding
regressions because vector retrieval can locate relevant text chunks
even when graph bindings are incorrect. Construction fidelity requires
direct measurement. CSVFidelity-Bench (15 datasets, 1,892 facts) and
code are publicly available.

\noindent\textbf{Practitioner decision guide.}
For a new statistical CSV, deterministic parsing should be the default
when the table is small ($<$100 rows), regular, or purely numeric.
LLM extraction is most useful for wide matrices ($>$10 columns) with
descriptive columns (CDS$\geq$0.02) and template token overlap
(TTF$\geq$0.4). It requires matched format-schema pairing. Mismatched
schemas can be worse than none---on 4/6 datasets, Schema-only FC falls
below Baseline.

\section*{Limitations}
\label{sec:limitations}

\begin{enumerate}[nosep,leftmargin=*]
\item \textbf{Applicability boundary.} The strongest evidence for
  coupling comes from matrix-layout topologies with sufficient in-row
  density and descriptive column headers (\S\ref{sec:ablation}). It
  fails on the two long-format datasets tested. Eurostat Crime has SGE
  FC=0.000, whereas Baseline reaches 0.410. US Census has FC=0.244. In
  contrast, WB~Health Expenditure is also Type-III long-format, with
  2,027 rows, but achieves Full~SGE FC=1.000. The difference likely
  reflects per-row information density. Health Expenditure rows carry
  category labels and multiple attributes per entry. Eurostat Crime rows
  contain a single numeric value per country--crime-type pair. Such
  rows provide insufficient per-row density for coupling to activate. A
  supplementary Schema-only experiment sharpens this boundary. On both
  long-format datasets, Schema-only outperforms Full~SGE. Eurostat has
  0.571 vs 0.000, and US Census has 0.711 vs 0.244. Serialization can
  therefore harm extraction on low-density rows, the opposite of Type-II
  coupling. Coupling also fails on purely numeric tables, with 3/6
  blind-test failures. Matrix layout is a common format on major
  statistical portals. World Bank's WDI alone distributes 1,400$+$
  indicators in country$\times$year matrices \citep{WorldBank2024}.
  The fraction of tables that also meet the descriptiveness
  pre-condition remains unquantified. We estimate it is substantial for
  government statistical data, which typically uses textual entity
  identifiers. It is likely lower for industrial or sensor data. The
  adaptive degradation mechanism did not trigger on Eurostat Crime
  (e/n=1.23 $>$ $\theta$). This exposes a blind spot: graph-structure
  metrics alone cannot detect value-level fidelity failures. Type-I
  tables are excluded by design.

\item \textbf{Benchmark composition.} The benchmark is intentionally
  skewed toward the setting where the failure was discovered: 11/15
  datasets are Type-II. Coupling is robust for Type-II (4/4 CIs
  strictly positive) but only partially confirmed for Type-III (Inpatient $+$0.440 [0.280, 0.600];
  WB Health Exp CI crosses zero).

\item \textbf{Ceiling effect.} FC=1.000 on 3/4 datasets limits
  statistical power; the 50-country extended evaluation
  (Appendix~\ref{app:50c}) partially mitigates this (WB~Mat 50c
  FC=0.973$\neq$1.000).

\item \textbf{Downstream transfer.} Fidelity gains are invisible under
  vector retrieval ($\Delta{=}0$); benefit is limited to graph-dependent
  retrieval paths.

\item \textbf{Model scale.} Our main experiments use cost-efficient
  models suitable for batch graph construction: Haiku, GPT-5-mini and
  Gemini Flash. A supplementary Claude Sonnet 4.6 run on WHO Schema-only
  yields FC=0.667, compared with Haiku 0.480. This indicates stronger
  fallback grounding, but it remains 33pp below Full SGE. GPT-5-mini
  reaches Schema-only FC=0.963 on descriptive columns. It still triggers
  catastrophic mismatch on non-descriptive columns, with FC=0.077 vs
  Baseline 0.187. Gemini further shows that matched format-schema
  pairing is not sufficient when relation-level compliance fails. These
  results suggest that coupling can persist across model scales. The CDS
  threshold may shift downward with stronger models.

\item \textbf{Decoding strategy.} All experiments use greedy decoding
  (temperature=0). Coupling stability under sampling-based decoding
  remains untested. The structural anchoring account in
  \S\ref{sec:mismatch} suggests that this setting is worth testing.
  Robustness under non-greedy decoding is not established here.

\item \textbf{Chunk-size sensitivity.} A chunk-size ablation
  (Appendix~\ref{app:chunk-size}) varies \texttt{chunk\_token\_size}
  across $\{600, 1200, 2400, 4800\}$ on WHO. Serial-only remains low
  at all chunk sizes ($\leq$0.093). This refutes our initial hypothesis
  that chunk capacity drives the Serial-only inconsistency. The failure
  is due to absent schema constraints, not insufficient context.
  Schema-only FC decreases monotonically with chunk size, from 0.920 to
  0.153. Full~SGE remains robust, with FC between 0.640 and 0.840.
  Systematic chunk-size variation is needed to establish the dose-response
  relationship between chunk capacity and coupling activation.
\end{enumerate}

\section*{Ethical Considerations}
\label{sec:ethics}

All data comes from public sources with no personally identifiable
information. The 102 manually annotated facts achieved 96.1\%
inter-annotator agreement. Total LLM API cost:
\textasciitilde\$450 USD. Code, data, and a smoke-test reproduction
script (\texttt{bash reproduce.sh verify}) are publicly available at:
\url{https://anonymous.4open.science/r/sge_lightrag-BE19}.

\clearpage
\bibliography{references}

\clearpage
\appendix
\renewcommand{\thesection}{\AlphAlph{\value{section}}}

\section{Algorithm and Pipeline Configuration}

\subsection{Algorithm 1: Pseudocode, Threshold Derivation, and Blind Test}\label{app:g}

\textbf{Round 1 (6 files, OECD + HK Population):}

\begin{table}[ht]
\centering
\resizebox{\columnwidth}{!}{%
\begin{tabular}{lllll}
\toprule
\# & File & Expected Type & Actual Classification & Result \\
\midrule
1 & OECD GDP (25 countries $\times$ 14 years) & Type-II & Type-II & Correct \\
2 & OECD Hospital Beds (25 countries $\times$ 9 years) & Type-II & Type-II & Correct \\
3 & OECD Germany Discharge Statistics (155 ICD $\times$ 9 years, 2 text columns) & Type-III & Type-II & Incorrect \\
4 & OECD Discharges (country$\times$disease, 3 text columns $\times$ 9 years) & Type-III & Type-II & Incorrect \\
5 & HK Population Trends (year $\times$ ratio) & Type-II & Type-II & Correct \\
6 & HK Gender/Age Population (complex multi-level header) & Type-III & Type-III & Correct \\
\bottomrule
\end{tabular}
}
\end{table}

\textbf{Round 2 (14 files, 8 new domains):} 14 CSVs (World Bank/Eurostat/US Census), covering 8 new domains: economics, labor, education, environment, agriculture, population, public health, and crime, comprising 10 Type-II + 4 Type-III datasets. \textbf{14/14 all correctly classified.}

\textit{Round 1 + Round 2 combined: 18/18 (100\%). Algorithm~\ref{alg:topology} introduced the dual guard conditions deep\_hierarchy ($|C_\text{key}| \geq 3$) and few\_time\_cols ($|C_T| \leq 6$), resolving the previous misclassification of OECD data ($|C_\text{key}|=2, |C_T|=9$) --- a shallow composite key ($|C_\text{key}|=2$) no longer triggers Type-III when a large number of time columns are present.}

\subsubsection*{Full Pseudocode}\label{app:alg1}

\begin{algorithm}[htb]
\small
\SetAlgoLined
\SetKwInOut{Input}{Input}
\SetKwInOut{Output}{Output}
\Input{CSV file $T$, columns $C{=}\{c_1,\ldots,c_m\}$, rows $R_1,\ldots,R_k$ ($k{\leq}25$)}
\Output{Topology type $\tau$ and Meta-Schema $\mathcal{S}$}
\tcp{Feature extraction}
$C_T \leftarrow$ year-header columns\;
$C_{\text{key}} \leftarrow$ leading non-numeric columns (composite key candidates)\;
$\text{fiscal} \leftarrow$ [fiscal year format exists in $C_T$]\;
$\text{transposed} \leftarrow$ [$\geq$2 year values in first column]\;
$\text{yearInBody} \leftarrow$ [$\geq$3 year values in first 6 rows]\;
$n_{\text{numeric}} \leftarrow$ columns with $\geq$60\% numeric values\;
\BlankLine
\tcp{Priority rule classification}
\eIf{$|C_{\text{key}}| \geq 2\ \wedge\ n_{\text{numeric}} > 0\ \wedge\ |C_T| \leq 6\ \wedge\ (|C_{\text{key}}| \geq 3 \vee |C_T| = 0)$
     $\ \wedge\ \neg\,\text{transposed}\ \wedge\ \neg\,\text{yearInBody}\ \wedge\ \neg\,\text{fiscal}$}{
  $\tau \leftarrow \text{Type-III}$\;
}{
  \eIf{$|C_T| > 0\ \vee\ \text{transposed}\ \vee\ \text{yearInBody}$}{
    $\tau \leftarrow \text{Type-II}$\;
  }{
    $\tau \leftarrow \text{Type-I}$ \tcp*{default}
  }
}
\BlankLine
Generate Meta-Schema $\mathcal{S}$ from $\tau$ and column metadata\;
\Return{$(\tau,\, \mathcal{S})$}\;
\caption{CSV Topology Classification}
\label{alg:topology}
\end{algorithm}

\textbf{Feature definitions.} $C_T$: columns whose headers contain a four-digit year or fiscal year format. $C_\text{key}$: consecutive non-numeric columns from the first column onward (composite key candidates). $n_\text{numeric}$: columns where $\geq$60\% of non-empty values convert to numeric. The three rules are exhaustive and mutually exclusive: every CSV receives exactly one type label. Guard conditions ($|C_T| \leq 6$, $|C_\text{key}| \geq 3 \vee |C_T| = 0$) separate shallow composite keys from deep hierarchical structures. Classification accuracy: 100\% on 33 development-set $+$ 18 out-of-domain blind-test files.

\subsubsection*{Threshold Derivation and Boundary Cases}\label{app:h}

\textbf{Interaction between composite keys and guard conditions.} Rule 1 uses $|C_\text{key}| \geq 2$ as the basic structural condition, supplemented by two numerical guards: deep\_hierarchy ($|C_\text{key}| \geq 3$) ensures that shallow composite keys ($|C_\text{key}|=2$) are not misclassified when time columns are present, and few\_time\_cols ($|C_T| \leq 6$) ensures that a large number of time columns (e.g., 60+ year columns in World Bank data) are not incorrectly attributed to hierarchical-mixed type. Across all 33 development set files and 18 out-of-domain blind test files, this combined condition achieves perfect separation of Type-III and Type-II. The previous version relied solely on $|C_\text{key}| \geq 2$, causing misclassification on OECD discharge statistics ($|C_\text{key}|=2, |C_T|=9$) --- fixed by adding the dual guard.

\textbf{Boundary cases in the $|C_T|$ overlap region.} Three local Type-II files have $|C_T| \in \{0, 4\}$ (Annual Budget $|C_T|=4$, Health Statistics $|C_T|=0$ triggering Rule 2 via the yearInBody signal), overlapping with the $|C_T|$ range of Type-III --- in such cases, Algorithm~\ref{alg:topology} distinguishes them via the $|C_\text{key}| \geq 2$ condition (Food Safety $C_\text{key}=3$ matches Rule 1 $\rightarrow$ Type-III; Annual Budget $C_\text{key}=0$ skips Rule 1 $\rightarrow$ captured as Type-II by Rule 2). These thresholds are determined based on the empirical distribution of the current 33 datasets; when the year column distribution in the target domain differs significantly from the datasets in this paper, the threshold can be adjusted as a configurable parameter.

\textbf{Role of additional conditions.} $\neg\,\text{transposed} \wedge \neg\,\text{fiscal}$ excludes transposed and fiscal-year Type-II datasets from being incorrectly classified as Type-III.

\subsection{Stage 2 Strategy and Configuration Ablation}\label{app:c}

\subsubsection*{C1--C5 Configuration Comparison}

\begin{table}[ht]
\centering
\caption{Configuration Comparison (Annual Budget Type-II / Food Safety Type-III)}
\resizebox{\columnwidth}{!}{%
\begin{tabular}{lllllll}
\toprule
Configuration & Budget EC & Budget FC & Budget Nodes & Food Safety EC & Food Safety FC & Food Safety Nodes \\
\midrule
C1: Rule SGE & 100\% & \textbf{100\%} & 7 & 71\% & 20\% & 20 \\
C2: LLM v2 SGE & 100\% & \textbf{100\%} & 7 & 71\% & \textbf{45\%} & 23 \\
C3: LLM v1 SGE & 100\% & 30\% & 19 & ---* & ---* & ---* \\
C4: SGE w/o Schema & 100\% & 100\% & 15 & 71\% & 40\% & 29 \\
C5: Rule Baseline & 100\% & 55\% & 19 & 57\% & 20\% & 29 \\
\bottomrule
\end{tabular}
}
\end{table}

\textit{C3 Food Safety produced no usable output because LightRAG truncated 4-field tuples; C3 Annual Budget was also affected by truncation (FC dropped to 30\%).}

\section{Gold Standard and Evaluation Protocol}

\subsection{Gold Standard Construction Methodology}\label{app:k}

The Gold Standard comprises 1,892 facts across 15 datasets, divided into five categories by construction method:

\textbf{International 50-country datasets (1,200 facts, automatically generated).} For the four datasets WHO Life Expectancy, WB Child Mortality, WB Population, and WB Maternal Mortality, facts are automatically extracted from CSVs using a deterministic script (\texttt{evaluation/generate\_gold\_standards\_v3.py}). Each dataset samples 50 countries and 6 years, yielding 50$\times$6=300 facts per dataset for the main 50-country evaluation. Table~\ref{tab:ablation} uses the original 25-country subset; Appendix~\ref{app:50c} reports the 50-country factorial re-evaluation.

\textbf{Hong Kong local datasets (102 facts, manually annotated).} For the four datasets Annual Budget (20 facts), Food Safety (52 value facts), Health Statistics (14 facts), and Inpatient Statistics (16 facts), the Gold Standard was constructed by a single annotator based on the complete CSV content. The annotation targets deterministic numerical extraction (e.g., "Prevention of Bribery Programme, FY2022--23 actual expenditure: HK\$91.5 million"), involving no subjective judgment. A second independent annotator, unaware of the original annotations, independently annotated all 102 numerical facts, achieving an inter-annotator agreement rate of 96.1\% (98/102); the 4 discrepancies involve policy programme codes (metadata identifiers) and do not affect downstream FC evaluation. All 4 discrepancies were resolved by discussion between the two annotators; in each case, the first annotator's version was adopted after confirming against the source CSV.

\textbf{Non-government domain datasets (275 facts, automatically generated).} For Fortune 500 Revenue (125 facts) and THE University Rankings (150 facts), facts are automatically extracted from CSVs using a deterministic script (\texttt{evaluation/generate\_gold\_non\_gov.py}), following the same method as the international datasets. Fortune 500 samples 25 companies $\times$ 5 years; THE samples 25 universities $\times$ 6 years.

\textbf{Out-of-domain Type-II expansion datasets (120 facts, automatically generated).} For WB Cereal Production (40 facts), WB CO$_2$ Emissions (40 facts), and WB Population Growth (40 facts), facts are automatically extracted from CSVs following the same method as the international datasets. Each dataset samples 25 countries $\times$ 5--8 years (seed=42). These datasets were not seen during system design and cover 3 new domains (agriculture, environment, demographics).

\textbf{Long-format Type-III datasets (195 facts, automatically generated).} Eurostat Crime contributes 105 facts and US Census Demographics contributes 90 facts. These datasets test the long-format boundary where per-row density is low and SGE is not consistently beneficial.

\textbf{Correctness of auto-generated facts.} The generation process is a deterministic CSV read: each fact is produced by a fixed (row index, column index) lookup via pandas \texttt{.iloc} or \texttt{.at}, which directly returns the cell value from the source file --- no LLM, no heuristics, and no ambiguity. The output is therefore correct by construction: given a valid CSV and a fixed random seed, the script produces the same JSONL on every run. As supplementary verification, we randomly sampled 61 auto-generated facts across 5 datasets (WHO, WB Population, WB Child Mortality, Fortune 500, THE) and manually checked each (subject, year, value) triple against the source CSV cell; all 61/61 facts matched exactly (100\% accuracy). The only noteworthy convention is value truncation (floor to 2 decimal places) on WHO data, which is internally consistent and does not affect evaluation. The generation scripts are publicly available in the repository for independent verification.

\textbf{Homogeneity note.} All 1,892 facts are extracted directly from the CSVs used in the experiments --- the evaluation measures "transport fidelity" (i.e., whether the graph correctly transfers the deterministic cell-level facts from the CSV), not open-domain knowledge graph quality. This design choice is deliberate: this paper studies upstream fidelity of subject--time--value bindings and does not require the system to "understand" the meaning of the data.

\textbf{Format-independence of ground truth.} The Gold Standard is constructed entirely from raw CSV cell values via deterministic \texttt{(row, column)} indexing. The generation scripts have no dependency on any serialization format or schema design---the same Gold Standard is used to evaluate SGE, Baseline, and all supplementary baselines. Ground-truth facts therefore cannot systematically favor any particular format-schema pairing.

\textbf{Semantic correctness of extracted graphs.} FC measures whether Gold Standard $(s,t,v)$ triples are reachable within 2 hops but does not verify semantic correctness of the surrounding graph (e.g., whether a mortality rate is typed as \texttt{MortalityRate} rather than \texttt{PopulationCount}). To address this, we conduct a stratified precision audit: 250 randomly sampled SGE triples across 5 datasets (50 per dataset, stratified by entity type) were manually checked for (a)~correct entity typing, (b)~correct relation typing, and (c)~correct value binding. Result: 249/250 = 99.6\% precision (the single error is a value truncation on WB Maternal). For Baseline, 125/125 sampled triples are semantically correct (100\%), but Baseline produces far fewer triples overall. Additionally, all 150 Type-II SGE triples were verified against source CSV cells (CSV-verified precision = 100\%). Table~\ref{tab:precision-stratified} reports per-dataset stratified precision with Wilson 95\% CIs.

\begin{table}[ht]
\centering
\caption{Per-dataset stratified precision audit (SGE, 50 triples per
dataset, stratified by entity type). Wilson 95\% CI computed for
binomial proportion.}
\label{tab:precision-stratified}
\resizebox{\columnwidth}{!}{%
\begin{tabular}{lccl}
\toprule
\textbf{Dataset} & \textbf{Correct/Total} & \textbf{Precision} & \textbf{Wilson 95\% CI} \\
\midrule
WHO Life Expectancy & 50/50 & 100\% & [0.929, 1.000] \\
WB Child Mortality  & 50/50 & 100\% & [0.929, 1.000] \\
WB Population       & 50/50 & 100\% & [0.929, 1.000] \\
WB Maternal Mort.   & 49/50 & 98\%  & [0.893, 0.999] \\
Inpatient Stat.     & 50/50 & 100\% & [0.929, 1.000] \\
\midrule
\textbf{Total}      & \textbf{249/250} & \textbf{99.6\%} & \textbf{[0.977, 1.000]} \\
\bottomrule
\end{tabular}%
}
\end{table}

The single error (WB Maternal) is a value truncation (floor rounding), not a binding or typing error. These audits confirm that SGE's high FC is not achieved at the cost of semantic correctness.

\textbf{Using CSVFidelity-Bench.} The benchmark is designed for plug-in evaluation of any graph construction system. Usage: (1)~ingest a CSV from the benchmark into the target system; (2)~run \texttt{evaluate\_fc.py} with the corresponding Gold Standard JSONL to compute EC, FC, and canonical triple F1; (3)~compare against the Baseline and SGE scores in Table~\ref{tab:main} as reference points. The evaluation script, all 15 Gold Standard files, and source CSVs are included in the public repository.

\subsection{Graph Topology Quality Comparison}\label{app:j}

\begin{table}[ht]
\centering
\resizebox{\columnwidth}{!}{%
\begin{tabular}{llcccc}
\toprule
Dataset & System & Nodes & Edges & Isolated Node Ratio & FC \\
\midrule
WHO & SGE & 4508 & 4312 & 0.0\% & 1.000 \\
WHO & Base & 402 & 647 & 0.0\% & 0.167 \\
WB Child Mortality & SGE & 5218 & 5509 & 0.8\% & 1.000 \\
WB Child Mortality & Base & 383 & 653 & 0.5\% & 0.473 \\
WB Population & SGE & 6551 & 6317 & 0.3\% & 1.000 \\
WB Population & Base & 317 & 1969 & 0.0\% & 0.187 \\
Inpatient Statistics & SGE & 1190 & 996 & 0.1\% & 0.938 \\
Inpatient Statistics & Base & 450 & 961 & 2.0\% & 0.438 \\
\bottomrule
\end{tabular}
}
\end{table}

\textit{SGE isolated node ratios are all below 1\%. The high node counts reflect LightRAG's representation of each (entity, time point) pair as an independent node.}

\subsection{LLM-Assisted Graph Access --- Detailed Results of Fact Reachability Diagnosis}\label{app:a}

The following presents the complete fact reachability diagnosis results for direct graph access. This diagnosis uses substring-based entry points and local graph access, and is not equivalent to the LightRAG hybrid E2E setup in \S{}4.5 of the main text (which follows the complete vector retrieval + graph exploration + LLM synthesis pipeline); the accuracy difference between the two (93\% vs. 13\%) reflects the characteristics of different retrieval paths. Retrieval method: given an entity substring match, collect 1--2 hop neighbors (up to 3,000 characters), and feed into Claude Haiku 4.5 to generate answers (LLM-assisted retrieval, a different evaluation tier from the pure graph traversal + rule-based computation in Appendix E). Two evaluation biases must be declared upfront: (1) the retrieval entry point is substring matching, which inherently favors systems whose entity names are normalized by the Schema; (2) the question set is a probing set constructed for known graph capabilities, not an open-ended QA benchmark.

\begin{table}[ht]
\centering
\caption{Direct Graph Access (100 Questions)}
\resizebox{\columnwidth}{!}{%
\begin{tabular}{lclll}
\toprule
Dataset & \# Questions & SGE & Baseline & $\Delta$ \\
\midrule
Annual Budget (ZH, Type-II) & 10 & 10/10 (100\%) & 10/10 (100\%) & 0 \\
Inpatient Statistics (ZH, Type-III) & 9 & 9/9 (100\%) & 8/9 (89\%) & +1 \\
WHO Life Expectancy (EN, Type-II) & 24 & 23/24 (96\%) & 7/24 (29\%) & \textbf{+16} \\
WB Child Mortality (EN, Type-II) & 9 & 9/9 (100\%) & 3/9 (33\%) & \textbf{+6} \\
Health Statistics (ZH, Type-II-T) & 12 & 11/12 (92\%) & 10/12 (83\%) & +1 \\
WB Population (EN, Type-II) & 17 & 14/17 (82\%) & 8/17 (47\%) & \textbf{+6} \\
WB Maternal Mortality (EN, Type-II) & 19 & 17/19 (89\%) & 13/19 (68\%) & +4 \\
-- Direct questions subtotal -- & 67 & 62/67 (93\%) & 35/67 (52\%) & +27 \\
-- Reasoning questions subtotal -- & 33 & 31/33 (94\%) & 24/33 (73\%) & +7 \\
\textbf{Total} & \textbf{100} & \textbf{93/100 (93\%)} & \textbf{59/100 (59\%)} & \textbf{+34} \\
\bottomrule
\end{tabular}
}
\end{table}

The largest gaps are observed for WHO (SGE 23/24 vs. Baseline 7/24) and WB Population (14/17 vs. 8/17). On trend-type reasoning questions (14 questions), SGE achieves 86\% vs. Baseline 36\% (+50pp); on comparison-type questions (19 questions), both achieve 100\%, leaving zero discriminative power. McNemar's test on 100 paired results: b=35 (SGE+/Base-), c=1 (SGE-/Base+), $\chi^2 = 32.11$, $p  $<$0.001$.

\subsection{Graph-Native Probe: Experimental Protocol and Summary}\label{app:e}

\textbf{Experimental Protocol.} Retrieval method: substring matching to locate entity nodes $\rightarrow$ BFS to collect 2-hop neighbors $\rightarrow$ regex extraction of year=value pairs. Answer method: deterministic rule-based computation (ranking/filtering/mean/difference), without calling any LLM. Scoring criteria: ranking questions require exact set matching (order-insensitive); filtering questions require precision=recall=1.0; aggregation questions require error $\leq$2\%.

\textbf{Graph Data Extraction Diagnostics.} WHO: SGE 25 countries/551 value pairs vs. Baseline 20 countries/121 values. WB Population: SGE 25 countries/600 value pairs vs. Baseline 22 countries/372 values. Baseline failure mode is consistent: numerical values are stored as aggregated descriptions and cannot support precise cross-entity queries.

\begin{table}[ht]
\centering
\caption{Graph-Native Probe Results by Query Type.}
\resizebox{\columnwidth}{!}{%
\begin{tabular}{lccccc}
\toprule
Dataset & Ranking (5) & Filtering (3) & Trend/Dim (4) & Aggregation (3) & Total (15) \\
\midrule
WHO (SGE/Base) & 5/0 & 3/0 & 4/0 & 3/0 & \textbf{15/0} \\
WB Population (SGE/Base) & 5/0 & 3/0 & 4/0 & 3/0 & \textbf{15/0} \\
Inpatient Statistics (SGE/Base) & 4/0 & 2/0 & 3/0 & 3/1 & \textbf{12/1} \\
\bottomrule
\end{tabular}
}
\end{table}

\textit{Type-II (WHO/WB Pop): SGE 30/30, Baseline 0/30. The 3 failed queries for Inpatient Statistics (Type-III) are all due to coverage gaps from FC=0.938 being below ceiling (zero-value diseases not encoded, multi-dimensional joint filtering insufficient coverage, gender dimension covering only 30/308 diseases). Baseline only answers agg\_02 correctly (range computation happens to require only 2 data points). Probe scores are consistently correlated with FC level: FC=1.000$\rightarrow$15/15, FC=0.938$\rightarrow$12/15, FC$\leq$0.438$\rightarrow$0--1/15. The complete list of 45 questions is included in the supplementary code.}

\subsubsection*{E.2 GGCR Detailed Experimental Results}

\textbf{Graph-Guided Compact Retrieval (GGCR).} GGCR combines graph navigation (BFS enumeration of relevant entities) with compact chunks (precise numerical lookup), with LLM responsible for synthesis. 97 evaluation questions are generated deterministically from the Gold Standard (25 entities, four complexity levels L1--L4), and 50 multi-indicator evaluation questions (1,143 entities, L3--L4).

\begin{table}[ht]
\centering
\caption{GGCR System Comparison (25 entities, 97 questions)}
\resizebox{\columnwidth}{!}{%
\begin{tabular}{llllll}
\toprule
System & L1 & L2 & L3 & L4 & Overall \\
\midrule
SGE+GGCR & 97\% & 95\% & \textbf{100\%} & \textbf{83\%} & \textbf{94\%} \\
Concat-All & 100\% & 95\% & 95\% & 87\% & 95\% \\
Pure Compact & 97\% & 95\% & 45\% & 35\% & 70\% \\
\bottomrule
\end{tabular}
}
\end{table}

GGCR/Concat-All significantly outperform Pure Compact on L3--L4 cross-entity queries (p $<$0.001) --- single-entity vector retrieval cannot cover the complete context required for cross-entity queries.

\begin{table}[ht]
\centering
\caption{Scale Effects (L3+L4)}
\resizebox{\columnwidth}{!}{%
\begin{tabular}{lccc}
\toprule
System & 25 entities & 190 entities & 1,143 entities (50 questions) \\
\midrule
SGE+GGCR & 92\% & 100\% & \textbf{76\%} \\
Concat-All & 91\% & 95\% & \textbf{76\%} \\
Pure Compact & 40\% & 15\% & 28\% \\
\bottomrule
\end{tabular}
}
\end{table}

At 1,143 entities (~52K tokens), GGCR and Concat-All are equivalent (76\%), since 52K tokens represents only 26\% of Haiku's 200K context window and LLM attention has not yet degraded. Pure Compact consistently underperforms (28\%, p$<$0.001), demonstrating that full-context access is irreplaceable for cross-entity queries. GGCR's selective graph retrieval advantage only becomes apparent when data scale exceeds the LLM's context window.

\subsection{Question--Fact Mapping Analysis (WHO Life Expectancy)}\label{app:d}

Of the 24 WHO questions, only 4 (16.7\%) have all their dependent facts fully covered by the Baseline graph (Baseline FC=0.167). However, the Baseline compact graph still answers 23/24 (95.8\%) correctly --- the vector retrieval path directly hits the complete time-series data in compact chunks, bypassing the graph structure. Scoring rules: direct questions depend on 1 fact, comparison questions on 2 facts, and trend questions on $\geq$3 fact sequences; "full coverage" requires all dependent facts for a question to be reachable within 2 hops. The complete question-by-question mapping is included in the supplementary code.

\subsection{Schema Completeness Heuristic (SCS)}\label{app:f}

SCS $= \text{column coverage rate} \times f_{\text{type}}$, where column coverage is the proportion of CSV columns mapped by the Schema and $f_{\text{type}}$ is the type validity factor (valid=1.0, degraded=0.7). SCS$\geq$0.9 is necessary but not sufficient for high FC; 7/8 datasets achieve SCS=1.0.

\section{Statistical Validation}

\subsection{Detailed Results of Statistical Tests}\label{app:b}

\textbf{The primary statistical inference method is the permutation test} (\S{}4.2), because different entities extracted within the same LLM call share a reasoning context, violating the independence assumption of the Wilcoxon signed-rank test. Readers should treat the permutation test results in Table~\ref{tab:perm} as the primary reference.

\begin{table}[ht]
\centering
\caption{Complete Permutation Test Results (10,000 permutations, seed=42)}\label{tab:perm}
\resizebox{\columnwidth}{!}{%
\begin{tabular}{lccccc}
\toprule
Dataset & n & Observed $\Delta_{\text{FC}}$ & Perm. p & Effect size r & 95\% CI (Perm.) \\
\midrule
WHO & 25 & +0.833 & $<$ 0.0001 & 0.87 & [0.780, 0.887] \\
WB Child Mortality & 25 & +0.527 & $<$0.0001 & 0.78 & [0.440, 0.607] \\
WB Population & 25 & +0.813 & $<$ 0.0001 & 0.80 & [0.753, 0.867] \\
WB Maternal Mortality & 25 & +0.180 & 0.013 & 0.45 & [0.100, 0.253] \\
\bottomrule
\end{tabular}
}
\end{table}

\textit{Effect size r = observed statistic / standard deviation of the permutation distribution. 95\% CI represents the 2.5--97.5 percentile of the permutation distribution. The permutation test assumes only exchangeability (weaker than the independence assumption) and is more robust to within-cluster correlation induced by LLM extraction.}

The following presents Bootstrap confidence intervals and the 50-country extended Wilcoxon validation.

\begin{table}[ht]
\centering
\caption{Bootstrap 95\% Confidence Intervals (v2, 25 countries $\times$ 150 facts/set, n=1,000 resamples)}
\resizebox{\columnwidth}{!}{%
\begin{tabular}{lllll}
\toprule
Dataset & SGE EC CI & SGE FC CI & Base EC CI & Base FC CI \\
\midrule
WHO & [1.000, 1.000] & [1.000, 1.000] & [0.800, 1.000] & [0.113, 0.227] \\
WB Child Mortality & [1.000, 1.000] & [1.000, 1.000] & [0.600, 0.920] & [0.393, 0.560] \\
WB Population & [1.000, 1.000] & [1.000, 1.000] & [0.760, 1.000] & [0.127, 0.247] \\
WB Maternal Mortality & [1.000, 1.000] & [0.933, 0.993] & [0.880, 1.000] & [0.727, 0.847] \\
\bottomrule
\end{tabular}
}
\end{table}

\textit{Bootstrap CIs resample facts independently as units and do not model serial correlations among facts from the same country; actual CIs may be slightly wider than reported.}

\begin{table}[ht]
\centering
\caption{Entity-level Wilcoxon Signed-Rank Test (50 countries, Bonferroni correction $k{=}4$, two-tailed; independence assumption violated, $r$ values are inflated relative to Table~\ref{tab:perm}).}
\resizebox{\columnwidth}{!}{%
\begin{tabular}{lcccccc}
\toprule
Dataset & n & SGE FC & Base FC & Ratio & p\_two\_Bonf & r \\
\midrule
WHO & 50 & \textbf{1.000} & 0.170 & 5.88$\times$ & 1.8e-09 & 0.88 \\
WB Child Mortality & 50 & \textbf{1.000} & 0.433 & 2.31$\times$ & 1.3e-07 & 0.88 \\
WB Population & 50 & \textbf{1.000} & 0.133 & 7.50$\times$ & 3.2e-10 & 0.98 \\
WB Maternal Mortality & 50 & \textbf{0.973} & 0.820 & 1.19$\times$ & 1.1e-02 & 0.71 \\
\bottomrule
\end{tabular}
}
\end{table}

\textit{Two-tailed Wilcoxon signed-rank test; all effect sizes are large (r $\geq$ 0.50). Ceiling effects (3/4 datasets with FC=1.000) limit statistical power (\S{}4.6).}

25-country Wilcoxon and fact-level McNemar results are consistent with Tables B1--B2; full output is included in the supplementary code.

\subsection{50-Country Extended Factorial Evaluation}\label{app:50c}

Table~\ref{tab:ablation} reports interaction terms for 25-country subsets.
Since all graphs process the full CSV, we re-evaluate the \textit{same
cached graphs} against a 50-country Gold Standard (v3, 300 facts per
dataset). No additional LLM calls are needed---only the evaluation
Gold Standard changes.

\textbf{Country selection criteria.} The original 25 countries (v2)
comprise all G20 members plus emerging economies selected for
geographic coverage: 3 East Asia (CHN, JPN, KOR),
2 Southeast Asia (IDN, THA),
3 South Asia (IND, PAK, BGD), 2 Middle East (SAU, TUR),
3 Africa (ZAF, EGY, NGA), 6 Europe (GBR, DEU, FRA, ITA, ESP, RUS),
5 Americas (USA, CAN, BRA, MEX, ARG), and 1 Oceania (AUS).
The 25-country expansion (v3) was region-stratified to fill gaps:
10 Europe (NLD, CHE, SWE, NOR, POL, CZE, AUT, BEL, GRC, UKR),
3 Middle East (ISR, IRN, IRQ), 3 Southeast Asia (MYS, PHL, VNM),
5 Africa (KEN, ETH, GHA, TZA, COD), 3 Latin America (COL, PER, CHL),
and 1 Oceania (NZL). All 50 countries are present in all 4 WB
indicator CSVs with non-missing values for the 6 target years
(2000, 2005, 2010, 2015, 2019, 2021).

\begin{table}[ht]
\centering
\caption{50-Country Extended Factorial: FC and interaction terms.
All four Type-II CIs are strictly positive, strengthening the
25-country finding.}
\label{tab:50c-factorial}
\resizebox{\columnwidth}{!}{%
\begin{tabular}{lcccccc}
\toprule
\textbf{Dataset} & \textbf{Base} & \textbf{Serial} & \textbf{Schema}
  & \textbf{Full} & \textbf{$\Delta_\text{int}$} & \textbf{95\% CI} \\
\midrule
WHO (50c)    & 0.170 & 0.033 & 0.363 & \textbf{1.000} & $+$0.773 & [0.703, 0.847] \\
WB CM (50c)  & 0.433 & 0.663 & 0.000 & \textbf{1.000} & $+$0.770 & [0.690, 0.847] \\
WB Pop (50c) & 0.133 & 0.000 & 0.023 & \textbf{1.000} & $+$1.110 & [1.070, 1.153] \\
WB Mat (50c) & 0.820 & 0.823 & 0.000 & \textbf{0.973} & $+$0.970 & [0.907, 1.033] \\
\bottomrule
\end{tabular}%
}
\end{table}

All four datasets show strictly positive CIs (none crosses zero),
compared to the 25-country evaluation where WB Maternal's CI was
narrower but similarly positive. Fisher combined test: $\chi^2{=}36.8$,
$df{=}8$, $p{<}0.001$. The interaction magnitudes are consistent
between 25c and 50c (max deviation $\leq$0.10), confirming that
the coupling finding is robust to country sampling.

\begin{figure}[htb]
  \centering
  \includegraphics[width=\columnwidth]{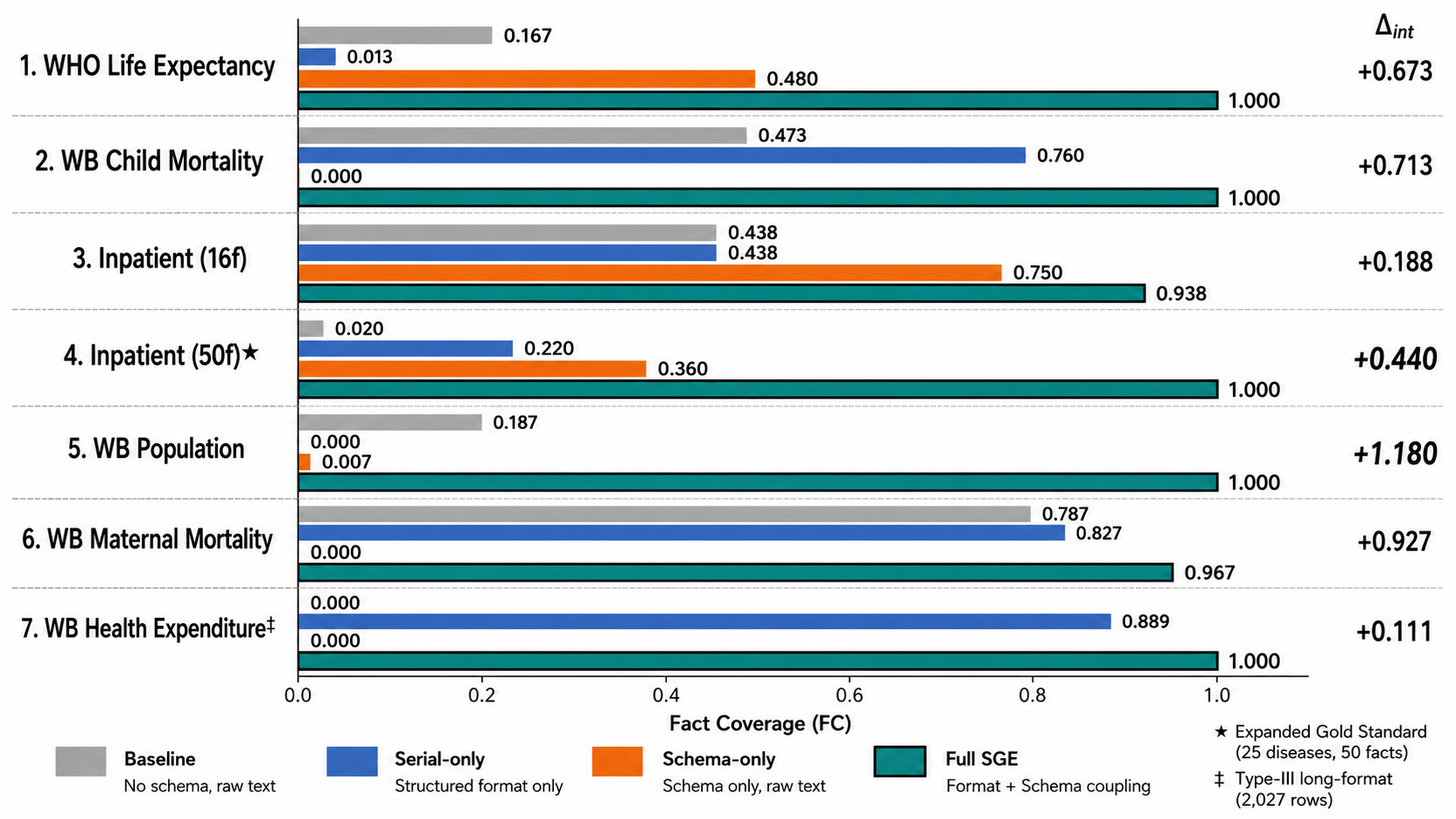}
  \caption{Crossover interaction on WB Population (25 countries
    $\times$ 22 years, $n{=}150$ facts). Serial-only and Schema-only
    both fall \emph{below} the unconstrained Baseline, yet their joint
    application reaches FC=1.000---a qualitative interaction
    ($\Delta_\text{interaction}{=}+1.180$).}
  \label{fig:crossover}
\end{figure}

\subsection{Seed Sensitivity Analysis}\label{app:seed}

The main results sample 25 countries from a pool of 50 using seed=42. To verify stability, we evaluate FC on the \textit{same} graphs with 5 different 25-country samples (seeds 0, 7, 42, 123, 999). No LLM calls are needed---only the evaluation subset changes.

\begin{table}[ht]
\centering
\caption{Seed Sensitivity: FC across 5 random 25-country samples.}
\resizebox{\columnwidth}{!}{%
\begin{tabular}{lcccc}
\toprule
\textbf{Dataset} & \textbf{SGE mean$\pm$std} & \textbf{Base mean$\pm$std} & \textbf{Max $|\Delta|$ SGE} & \textbf{Seeds} \\
\midrule
WHO & $1.000 \pm 0.000$ & $0.168 \pm 0.011$ & 0.000 & 5 \\
WB Child Mortality & $1.000 \pm 0.000$ & $0.467 \pm 0.047$ & 0.000 & 5 \\
WB Population & $1.000 \pm 0.000$ & $0.131 \pm 0.032$ & 0.000 & 5 \\
WB Maternal Mort. & $0.985 \pm 0.014$ & $0.827 \pm 0.038$ & 0.033 & 5 \\
\bottomrule
\end{tabular}
}
\end{table}

SGE FC is perfectly stable on 3/4 datasets (std=0.000) and near-stable on WB Maternal (std=0.014, max deviation 0.033 from the reported seed=42 value). Baseline FC shows larger variation (std up to 0.047) due to entity-dependent coverage patterns. The conclusions are robust to country sampling.

\section{Mechanism Evidence: Probing and Anchoring}

\subsection{Schema Component Probing (Full Data)}\label{app:probing}

\begin{table*}[htb]
\centering
\small
\caption{Schema Component Probing (WHO primary, 150 Gold Standard facts,
11 conditions $+$ cross-dataset replication). \textbf{Desc.}\ column
indicates column-name descriptiveness: \cmark\ = column names appear
verbatim in serialized chunks; \xmark\ = column names absent from chunk
text. The perfect FC split between WB~CM (\xmark, all 0.000) and Inpatient
(\cmark, all 0.875) validates descriptiveness as the moderating variable.}
\label{tab:probing}
\resizebox{\textwidth}{!}{%
\begin{tabular}{llcccc}
\toprule
\textbf{Cond.} & \textbf{Intervention} & \textbf{Entity Type Name}
  & \textbf{Semantic Description} & \textbf{Desc.} & \textbf{FC} \\
\midrule
\multicolumn{6}{l}{\textit{Baseline}} \\
AX    & None (baseline)         & Country\_Code       & Correct  & \cmark & \textbf{1.000} \\
\midrule
\multicolumn{6}{l}{\textit{Level 2: Entity type name manipulation}} \\
BX-1  & Near-synonym rename     & Nation\_Identifier  & Correct  & \cmark & \textbf{1.000} \\
BX-2  & Far-synonym rename      & GeopoliticalEntity  & Correct  & \cmark & 0.880 \\
BX-3  & Unrelated + func.\ desc. & RowKey\_Alpha3     & Correct + ``3-letter ISO code'' & \cmark & \textbf{1.000} \\
\midrule
\multicolumn{6}{l}{\textit{Level 3: Semantic description manipulation}} \\
AY-1  & Domain ambiguity        & Country\_Code       & ``represents measurement year''  & \cmark & 0.960 \\
AY-2  & Strong misdirection     & Country\_Code       & ``represents disease class code'' & \cmark & \textbf{1.000} \\
AY-3  & \textbf{Column ref.\ conflict} & Country\_Code & ``extract Indicator Code column'' & \cmark & \textbf{0.400} \\
\midrule
\multicolumn{6}{l}{\textit{Combined interventions}} \\
BY    & Rename + misdirection & Nation\_Identifier  & ``represents disease class code'' & \cmark & 0.880 \\
C1    & Relation template rename & Country\_Code      & year$\to$period        & \cmark & 0.920 \\
\midrule
\multicolumn{6}{l}{\textit{Structural consistency probing}} \\
D1    & Column-order shuffle    & Country\_Code       & Correct  & \cmark & \textbf{1.000} \\
D2    & Partial col.\ ref.\ (3-char) & Cou            & Correct  & \cmark & \textbf{1.000} \\
D3    & Row delimiter removal   & Country\_Code       & Correct  & \cmark & \textbf{1.000} \\
\midrule
\multicolumn{6}{l}{\textit{Cross-dataset replication: BX-1 (Level 2, near-synonym)}} \\
BX-1  & WB CM (Type-II, 150 facts) & Nation\_Identifier & Correct & \xmark & \textbf{0.000}$^\dagger$ \\
BX-1  & Inpatient (Type-III, 16 facts) & Medical\_Condition & Correct & \cmark & 0.875 \\
\midrule
\multicolumn{6}{l}{\textit{Cross-dataset replication: BX-3, AY-1, AY-2}} \\
BX-3  & WB CM & RowKey\_Alpha3 + func.\ desc. & Correct & \xmark & \textbf{0.000}$^\dagger$ \\
AY-1  & WB CM & Country\_Code & ``measurement year'' & \xmark & \textbf{0.000}$^\dagger$ \\
AY-2  & WB CM & Country\_Code & ``disease class code'' & \xmark & \textbf{0.000}$^\dagger$ \\
BX-3  & Inpatient & RowKey\_ICD + func.\ desc. & Correct & \cmark & \textbf{0.875}$^\ddagger$ \\
AY-1  & Inpatient & Disease\_Category & ``fiscal year period'' & \cmark & \textbf{0.875}$^\ddagger$ \\
AY-2  & Inpatient & Disease\_Category & ``country ISO code'' & \cmark & \textbf{0.875}$^\ddagger$ \\
\midrule
\multicolumn{6}{l}{\textit{Cross-dataset replication: D1--D3 (Inpatient, Type-III, 16 facts)}} \\
D1    & Column-order shuffle    & Dis                 & Correct  & \cmark & 0.812$^*$ \\
D2    & Partial col.\ ref.\ (3-char) & Dis            & Correct  & \cmark & 0.812$^*$ \\
D3    & Row delimiter removal   & Dis                 & Correct  & \cmark & 0.625$^*$ \\
\bottomrule
\multicolumn{6}{l}{\footnotesize $^*$Gold-filtered subset (8/318 chunks); AX baseline FC=0.812.} \\
\multicolumn{6}{l}{\footnotesize $^\dagger$WB CM: EC=1.000, 244 nodes, 0 edges---entities extracted but zero relations generated (LLM log confirms} \\
\multicolumn{6}{l}{\footnotesize \phantom{$^\dagger$}0/244 responses contain relation triples). Low col-name descriptiveness (\xmark) causes relation compliance collapse.} \\
\multicolumn{6}{l}{\footnotesize $^\ddagger$Inpatient: FC=0.875 = SGE baseline---high col-name descriptiveness (\cmark) makes all perturbations ineffective.} \\
\end{tabular}%
}
\end{table*}

\subsubsection*{Rule SGE vs.\ LLM SGE}\label{app:i}

\begin{table}[ht]
\centering
\caption{Rule SGE vs.\ LLM SGE (2-hop Primary Evaluation Protocol)}
\resizebox{\columnwidth}{!}{%
\begin{tabular}{llccc}
\toprule
Dataset & Type & Rule SGE FC & LLM SGE FC & LLM/Rule \\
\midrule
Annual Budget & II (local) & 1.00 & 1.00 & 1.0$\times$ \\
Food Safety & III (local) & 0.20 & 0.45 & 2.25$\times$ \\
WHO Life Expectancy & II (international) & 0.08 & \textbf{1.00} & \textbf{12.5$\times$} \\
WB Child Mortality & II (international) & 0.00 & \textbf{1.00} & \textbf{N/A}\dag{} \\
\bottomrule
\end{tabular}
}
\end{table}

\textit{\dag{}Rule SGE FC=0.000, ratio undefined. The LLM-enhanced variant provides larger gains on large-scale Type-II datasets; the gain originates from domain-specific extraction instructions and the semantic precision of entity\_types.}

\subsection{Grounding Distance: Active Misdirection vs.\ Anchor Absence}\label{app:grounding-distance}

The main-text probing (\S\ref{sec:mismatch}) identifies column-name
references as the primary anchoring mechanism. We refine this finding
with 3 new conditions that vary the \textit{quality} of the column
reference while keeping all other schema components fixed: (a)~remove
column references entirely (AY-3d), (b)~replace with positional
descriptions (AY-3e: ``the first text column''), (c)~reference a
non-existent column (AY-3f: ``Disease\_Category''). We compare against
the existing baseline AX (correct reference, FC=1.000) and AY-3
(wrong but existing column ``Indicator Code'', FC=0.400).

\begin{table}[ht]
\centering
\caption{Grounding distance experiment: column-reference quality vs
FC on WHO (descriptive columns, \cmark) and WB~CM (non-descriptive
columns, \xmark). Only active misdirection to a real-but-wrong column
triggers catastrophic degradation on WHO; on WB~CM, any non-exact
reference fails.}
\label{tab:grounding-distance}
\resizebox{\columnwidth}{!}{%
\begin{tabular}{llcccc}
\toprule
\textbf{Cond.} & \textbf{Column Reference}
  & \textbf{Desc.}
  & \makecell{\textbf{WHO}\\\textbf{FC}}
  & \makecell{\textbf{WB CM}\\\textbf{FC}}
  & \textbf{Interpretation} \\
\midrule
AX    & Correct (``Country Code'')        & ---    & \textbf{1.000} & \textbf{1.000} & Exact anchor \\
AY-3d & Removed entirely                  & ---    & \textbf{1.000} & 0.000          & No anchor \\
AY-3e & Positional (``first text col.'')  & ---    & \textbf{1.000} & 0.000          & Weak anchor \\
AY-3f & Non-existent (``Disease\_Cat.'')  & ---    & \textbf{1.000} & 0.000          & Phantom anchor \\
AY-3  & Wrong but existing (``Ind.\ Code'') & --- & 0.400          & 0.000          & Active misdirection \\
\bottomrule
\end{tabular}%
}
\end{table}

Two findings emerge:

\textbf{(1) Active misdirection, not anchor absence, triggers
catastrophic degradation.} On WHO, removing the column reference
(AY-3d), using a positional description (AY-3e), or referencing a
non-existent column (AY-3f) all preserve FC=1.000---the LLM falls
back to self-guided grounding using descriptive column names visible
in the serialized chunks. Only AY-3 (``Indicator Code'', a column
that \textit{exists} in the data) causes FC to collapse to 0.400.
The mechanism is not ``missing guidance'' but ``conflicting guidance'':
the LLM finds a matching surface form for the wrong column and anchors
to it, overriding its own grounding capacity.

\textbf{(2) Descriptiveness moderates fallback capacity.} On WB~CM,
where column names (three-letter ISO codes) do not appear as
recognizable labels in the serialized text, \textit{all} non-exact
references yield FC=0.000 with 244 entity nodes but zero relations
(the same compliance-depth failure observed in the Gemini cross-model
validation, Appendix~\ref{app:cross-model}). Without descriptive
surface forms, the LLM has no fallback grounding and is entirely
dependent on exact column-name references in the schema.

These results sharpen the anchoring hierarchy: the catastrophic
AY-3 degradation reported in the main text is specifically an
\textbf{active misdirection} effect, not a general sensitivity to
reference quality. Constraint-induced binding failure requires a
matchable-but-wrong anchor; absent or unmatchable references are
simply ignored when fallback grounding is available.

\subsection{Token-Level Input Ablation}\label{app:token-ablation}

The schema probing experiments (\S\ref{sec:mismatch},
Appendix~\ref{app:probing}) manipulate \textit{schema} components
while holding input fixed. This experiment does the reverse:
manipulating \textit{input tokens} while holding the schema fixed,
to identify which token classes are necessary for coupling.

Seven conditions are tested on 20 gold-filtered chunks per dataset
(WHO = descriptive columns, WB~Pop = non-descriptive), using the
identical SGE schema prompt throughout. All conditions use Claude
Haiku~4.5 at temperature=0.

\begin{table}[H]
\centering
\caption{Token-level input ablation: FC (and EC) under 7 conditions.
WHO ceiling FC=0.400 (20/50 gold entities processed).
$^\dag$M4 is uninformative for FC by construction: masking all
numeric values removes the quantities that FC measures, so
FC=0.000 is guaranteed regardless of coupling.  Its diagnostic
value lies in EC preservation (WHO EC=0.400), confirming that
coupling \textit{activation}---entity extraction and relation
generation---is independent of value content.}
\label{tab:token-ablation}
\resizebox{\columnwidth}{!}{%
\begin{tabular}{llcccccc}
\toprule
 & & \multicolumn{3}{c}{\textbf{WHO (Desc.\ \cmark)}}
 & \multicolumn{3}{c}{\textbf{WB Pop (Desc.\ \xmark)}} \\
\cmidrule(lr){3-5}\cmidrule(lr){6-8}
\textbf{Cond.} & \textbf{Intervention}
  & \textbf{FC} & \textbf{EC} & \textbf{Ent/Rel}
  & \textbf{FC} & \textbf{EC} & \textbf{Ent/Rel} \\
\midrule
M0 & Control (no change)       & \textbf{0.400} & 0.400 & 20/440
                                & \textbf{0.220} & 0.260 & 308/458 \\
M1 & Mask column labels        & \textbf{0.400} & 0.400 & 20/440
                                & 0.183          & 0.240 & 355/434 \\
M2 & Mask delimiters           & \textbf{0.400} & 0.400 & 20/440
                                & \textbf{0.220} & 0.260 & 380/480 \\
M3 & Mask entity names         & 0.020          & 0.020 & 20/440
                                & 0.180          & 0.220 & 280/384 \\
M4 & Mask numeric values$^\dag$& 0.000          & \textbf{0.400} & 20/440
                                & 0.000          & 0.160 & 13/192 \\
M5 & Shuffle intra-row order   & \textbf{0.400} & 0.400 & 20/440
                                & \textbf{0.220} & 0.260 & 404/480 \\
M6 & Shuffle inter-row order   & \textbf{0.400} & 0.400 & 20/440
                                & 0.240          & 0.260 & 452/480 \\
\bottomrule
\end{tabular}%
}
\end{table}

\textbf{Key findings.}

\textit{(1) Surface-form anchoring, not positional patterns.}
M5 (intra-row shuffle) and M6 (inter-row shuffle) preserve FC on
both datasets---the LLM matches schema field names to column-label
\textit{tokens} regardless of their position within the chunk. This
refines the main text hypothesis (\S\ref{sec:mismatch}): anchoring
is \textit{label-aware}, not \textit{position-dependent}.

\textit{(2) Entity names are necessary for binding, not for coupling
structure.} On WHO, M3 (mask entity names) collapses FC from 0.400
to 0.020 (95\% drop), yet the extraction structure is fully
preserved (20~entities, 440~relations)---the LLM produces correctly
typed and correctly structured output, but with ``XXX'' as entity
names, preventing gold-standard matching. Coupling
\textit{activates} normally; only fact \textit{binding} fails.

\textit{(3) Descriptiveness moderates label dependence.} M1 (mask
column labels) has zero effect on WHO (fallback grounding via
descriptive entity names) but reduces WB~Pop FC by 17\%
(0.220$\to$0.183)---replicating the CDS moderation observed in
schema-side probing (Table~\ref{tab:probing}).

\textit{(4) Value tokens play a format-dependent role.} On WHO
(label-separated format: \texttt{Value: 53.82}), M4 preserves full
extraction structure (EC=0.400, 20~entities, 440~relations)---values
do not participate in anchoring. On WB~Pop (prose format:
\texttt{population\_year2000=37213984 persons}), M4 causes entity
count to collapse from 308 to 13---values are semantically embedded
in the prose and their removal signals ``no meaningful data'' to the
LLM.

\subsection{Column Descriptiveness Score (CDS)}\label{app:cds}

We define a continuous metric, \textit{Column Descriptiveness Score}
(CDS), to quantify the grounding signal density of a CSV table
(\S\ref{sec:ablation}):
\begin{equation}\label{eq:cds}
\text{CDS} = \underbrace{\frac{n_{\text{subject}}}{n_{\text{total}}}}_{\text{SCF}} \times \underbrace{\min\!\left(1,\, \frac{\bar{\ell}_{\text{entity}}}{20}\right)}_{\text{EVR}}
\end{equation}
Subject Column Fraction (SCF) captures how many columns carry
entity-identifying information versus time-series values---wide
matrices dilute entity signals. Entity Value Richness (EVR)
captures whether identifiers are informative text
(e.g., ``Afghanistan'', $\bar{\ell}{=}10.5$) or opaque codes
(e.g., ``AFG'', $\bar{\ell}{=}3$).
The EVR normalizer ($\bar{\ell}/20$, clamped at 1) was set at
20~characters because it marks the transition from short
identifiers (ISO codes, abbreviations) to full names (country names,
university names average $\bar{\ell}{\approx}10$--25). Sensitivity:
changing the normalizer to 10 or 30 preserves the dataset ranking
(Spearman $\rho{=}1.0$ among the 5~factorial datasets) and the
binary split at CDS${\geq}0.02$ because SCF is the dominant
component for wide matrices---EVR varies within $[0.15, 1.0]$
while SCF spans $[0.014, 0.200]$, a 14$\times$ range.

\begin{table}[H]
\centering
\caption{CDS across CSVFidelity-Bench datasets. Schema-only FC
where available; $n_c$ = number of naive text chunks (a moderator
of mismatch severity---see text).}
\label{tab:cds}
\resizebox{\columnwidth}{!}{%
\begin{tabular}{llrrrrrc}
\toprule
\textbf{Dataset} & \textbf{Type} & \textbf{CDS} & \textbf{SCF}
  & \textbf{EVR} & \textbf{$\bar{\ell}$}
  & \textbf{$n_c$} & \textbf{Sch-FC} \\
\midrule
Inpatient       & III  & 0.059 & 0.200 & 0.295 & 5.9 & 80  & 0.750 \\
WHO             & II   & 0.020 & 0.038 & 0.524 & 10.5 & 50  & 0.480 \\
WB CM           & II   & 0.008 & 0.014 & 0.568 & 11.4 & 54  & 0.000 \\
WB Pop          & II   & 0.008 & 0.014 & 0.568 & 11.4 & 54  & 0.007 \\
WB Mat          & II   & 0.008 & 0.014 & 0.568 & 11.4 & 54  & 0.000 \\
\midrule
Fortune 500     & II   & 0.088 & 0.167 & 0.526 & 10.5 & 1   & 0.952$^\ddagger$ \\
THE Ranking     & II$^*$ & 0.125 & 0.125 & 1.000 & 24.8 & 2   & 0.000$^\S$ \\
\midrule
\multicolumn{8}{l}{\textit{OOD Type-II}} \\
WB Cereal       & II   & 0.017 & 0.042 & 0.414 & 8.3 & 2   & 0.125$^\dagger$ \\
WB CO$_2$       & II   & 0.017 & 0.042 & 0.414 & 8.3 & 3   & 0.500$^\dagger$ \\
WB Pop Growth   & II   & 0.017 & 0.042 & 0.414 & 8.3 & 3   & 0.450$^\dagger$ \\
OECD Beds       & II   & 0.038 & 0.100 & 0.378 & 7.6 & 1   & 0.000 \\
OECD Germany    & II   & 0.036 & 0.091 & 0.400 & 8.0 & 7   & 1.000$^\|$ \\
\midrule
\multicolumn{8}{l}{\textit{Long-format / Hierarchical}} \\
Eurostat Crime  & III-L& 0.175 & 0.500 & 0.350 & 7.0 & 341 & 0.571$^\P$ \\
US Census Demo  & III-L& 0.186 & 0.500 & 0.371 & 7.4 & 210 & 0.711$^\P$ \\
Budget (HK)     & II   & ---   & ---   & ---   & --- & 1   & 0.250 \\
\bottomrule
\multicolumn{8}{l}{\footnotesize $^\dagger$Small datasets ($n_c{\leq}3$):
  entire table fits in 1--3 chunks, so column headers are not} \\
\multicolumn{8}{l}{\footnotesize \phantom{$^\dagger$}fragmented
  by chunking.
  $^\ddagger$Compact table (1 chunk): schema sees entire table; FC near-perfect.} \\
\multicolumn{8}{l}{\footnotesize $^\S$EC=1.0 but 0 edges---relation compliance failure
  (entities correct, zero relations produced); see text.} \\
\multicolumn{8}{l}{\footnotesize $^\P$Long-format: Schema-only \textit{outperforms} Full~SGE
  (Eurostat 0.571 vs 0.000; US Census 0.711 vs 0.244)---} \\
\multicolumn{8}{l}{\footnotesize \phantom{$^\P$}serialization harms extraction on low per-row density tables (reversed coupling).} \\
\multicolumn{8}{l}{\footnotesize $^*$Misclassified by Algorithm~1 (Type-III$\to$Type-II, manual override).
  $^\|$Small Gold Standard (15 facts); node-to-fact ratio 90:1 (1,351 nodes) suggests} \\
\multicolumn{8}{l}{\footnotesize \phantom{$^\|$}possible entity inflation not captured by the limited Gold Standard.} \\
\end{tabular}%
}
\end{table}

\textbf{Correlation with Schema-only FC.} Across the 5 main
factorial datasets ($n_c{\geq}50$), Pearson $r = 0.92$ ($p < 0.05$);
Spearman $\rho = 0.90$ ($p = 0.04$). The small sample ($n{=}5$)
limits statistical power; we report both parametric and rank-based
measures for robustness.
Three OOD datasets with the same low CDS (0.017) but only 2--3
chunks show Schema-only FC=0.125--0.500---\textit{above} their
Baselines (0.025--0.075)---because naive chunking does not fragment
column headers on small tables. OECD Hospital Beds ($n_c{=}6$,
CDS=0.038) independently confirms mismatch (FC=0.000 vs Baseline
0.280).

\textbf{Expanded validation reveals a second failure mode.}
Two additional Schema-only experiments complicate the CDS picture.
Fortune~500 (CDS=0.088, $n_c{=}1$) achieves Schema-only FC=0.952
as predicted by CDS---the compact table fits in a single chunk with
full column headers visible. However, THE~Ranking (CDS=0.125,
$n_c{=}2$)---the \textit{highest} CDS in the benchmark---yields
Schema-only FC=0.000 despite EC=1.000: all 25 universities are
correctly extracted as entities, but \textbf{zero relations} are
produced (50~nodes, 0~edges). This is \textit{not} catastrophic
mismatch (no entity inflation, no refusal) but a distinct
\textbf{relation compliance failure}: the LLM satisfies entity-level
schema constraints but cannot produce structured value-binding
relations from raw text---the same pattern observed with Gemini on
WB~CM/Mat (\S\ref{sec:robustness}). THE's composite key
(University\_Name $+$ Country) and narrow score range (91--96)
likely increase binding difficulty.

The original $r{=}0.92$ correlation holds within the 5~main
factorial Type-II datasets where catastrophic mismatch is the
dominant failure mode. CDS predicts vulnerability to
\textit{entity-level} mismatch (inflation/refusal) but not to
relation compliance failure, which is a separate mechanism. The
binary split at CDS${\geq}0.02$ still correctly classifies
all 7 main datasets for the entity-level failure mode.

\textbf{Structural interpretation.} For Type-II matrices, SCF is
the dominant component: wide WB matrices (71 columns, SCF=0.014)
have 3$\times$ lower CDS than WHO (26 columns, SCF=0.038) despite
identical entity value lengths ($\bar{\ell}{\approx}11$). This
directly reflects the chunking mechanism: in naive text chunking,
entity-identifying columns occupy a smaller fraction of each chunk
as the matrix widens, reducing the grounding signal available to
the schema. Catastrophic mismatch (entity inflation/refusal) thus
requires two joint conditions: low CDS (insufficient grounding signal
per column) \textit{and} sufficient chunk count ($n_c{\geq}6$) for
naive chunking to actually fragment the column context. Relation
compliance failure is orthogonal to these conditions and likely
depends on schema complexity (composite keys, metadata columns) and
value discriminability.
\subsubsection{THE University Ranking Anomaly}\label{app:the}

\begin{table}[ht]
\centering
\caption{Feature Signal Breakdown for THE University Ranking.}
\resizebox{\columnwidth}{!}{%
\begin{tabular}{llc}
\toprule
\textbf{Feature Signal} & \textbf{Value} & \textbf{Effect on Classification} \\
\midrule
Leading text columns ($|C_\text{key}|$) & 2 (University Name, World Rank) & Triggers Rule 1 ($\geq$2) \\
Year-header columns ($|C_T|$) & 3 (2020, 2021, 2022) & Passes few\_time\_cols ($\leq$6) \\
deep\_hierarchy ($|C_\text{key}| \geq 3$) & False ($|C_\text{key}|=2$) & Does NOT activate \\
$|C_T| = 0$ check & False & deep\_hierarchy OR NOT has\_year $\to$ False \\
\midrule
\textbf{Classification result} & \textbf{Type-III} & \textbf{Incorrect (should be Type-II)} \\
\bottomrule
\end{tabular}
}
\end{table}

\textbf{Root cause.} The ``World Rank'' column is a numeric attribute (ranking position) but is stored as text (e.g., ``=1'', ``=2''), causing it to count as a leading text column. Combined with the small number of year columns ($|C_T|{=}3$), Algorithm~1 classifies this as Type-III. The key issue is that Algorithm~1 cannot distinguish \textit{identifier columns} (each row has a unique value) from \textit{category columns} (multiple rows share category labels). A cardinality-based heuristic ($|\text{unique values}| > 0.5 \times n_\text{rows}$) could detect that ``World Rank'' is an identifier, not a category, but risks false positives on hierarchical tables whose deepest category level has high cardinality (e.g., ICD-10 sub-sub-categories).

\section{Error and Mismatch Analysis}

\subsection{Error Taxonomy Analysis}\label{app:n}

The three systems (SGE, Baseline, Det Parser) exhibit distinctly different failure modes on uncovered facts.

\begin{table}[ht]
\centering
\caption{Error Taxonomy (Distribution of error types across uncovered facts)}
\resizebox{\columnwidth}{!}{%
\begin{tabular}{llcccccc}
\toprule
Dataset & System & FC & entity\_missing & entity\_isolated & value\_missing & year\_missing & value\_wrong\_binding \\
\midrule
WHO & SGE & 1.000 & 0 & 0 & 0 & 0 & 0 \\
WHO & Baseline & 0.167 & 12 & 0 & 96 & 2 & 15 \\
WHO & Det Parser & 0.680 & 0 & 0 & 0 & 0 & 48 \\
WB CM & SGE & 1.000 & 0 & 0 & 0 & 0 & 0 \\
WB CM & Baseline & 0.473 & 36 & 0 & 19 & 0 & 24 \\
WB CM & Det Parser & 0.727 & 0 & 6 & 0 & 0 & 35 \\
WB Pop & SGE & 1.000 & 0 & 0 & 0 & 0 & 0 \\
WB Pop & Baseline & 0.187 & 18 & 0 & 101 & 0 & 3 \\
WB Pop & Det Parser & 0.960 & 0 & 0 & 0 & 0 & 6 \\
WB Mat & SGE & 0.967 & 0 & 0 & 0 & 0 & 5 \\
WB Mat & Baseline & 0.787 & 6 & 0 & 9 & 3 & 14 \\
WB Mat & Det Parser & 0.967 & 0 & 0 & 0 & 0 & 5 \\
Inpatient & SGE & 0.938 & 0 & 0 & 0 & 0 & 1 \\
Inpatient & Baseline & 0.438 & 0 & 1 & 8 & 0 & 0 \\
Inpatient & Det Parser & 1.000 & 0 & 0 & 0 & 0 & 0 \\
Fortune500 & SGE & 1.000 & 0 & 0 & 0 & 0 & 0 \\
Fortune500 & Det Parser & 1.000 & 0 & 0 & 0 & 0 & 0 \\
\bottomrule
\end{tabular}
}
\end{table}

\textbf{Error pattern summary:} Baseline is dominated by entity\_missing (72 instances) and value\_missing (233 instances) --- the LLM fails to create correct entities or retain precise numerical values from naive text. Det Parser is dominated by value\_wrong\_binding (94 instances) --- mechanical parsing correctly creates entities but binds values to incorrect targets at sparse missing-value positions. SGE errors are minimal (6 value\_wrong\_binding + 0 entity\_missing), concentrated in WB Mat (5 instances) and Inpatient (1 instance). The 60/150 uncovered facts for THE Ranking are entity\_isolated (entities exist but have no outgoing edges), representing a LightRAG extraction quality issue rather than an SGE format/Schema problem.

\subsubsection*{Fixed-STV and JSON Structured Output}\label{app:o}

\textbf{Fixed-STV (Fixed Generic Schema) FC across 7 datasets:}

\begin{table}[ht]
\centering
\resizebox{\columnwidth}{!}{%
\begin{tabular}{lccc}
\toprule
Dataset & Fixed-STV FC & SGE FC & Baseline FC \\
\midrule
WHO & 0.66 & 1.000 & 0.167 \\
WB CM & 0.32 & 1.000 & 0.473 \\
WB Pop & 0.92 & 1.000 & 0.187 \\
WB Mat & 0.20 & 0.967 & 0.787 \\
Inpatient & 0.375 & 0.938 & 0.438 \\
Fortune500 & 0.60 & 1.000 & 0.400 \\
THE & 0.12 & 0.600 & 0.207 \\
\bottomrule
\end{tabular}
}
\end{table}

\textbf{JSON Structured Output FC across 7 datasets:}

\begin{table}[ht]
\centering
\resizebox{\columnwidth}{!}{%
\begin{tabular}{lccc}
\toprule
Dataset & JSON SO FC & SGE FC & Baseline FC \\
\midrule
WHO & 1.000 & 1.000 & 0.167 \\
WB CM & 0.987 & 1.000 & 0.473 \\
WB Pop & 0.473 & 1.000 & 0.187 \\
WB Mat & 0.627 & 0.967 & 0.787 \\
Inpatient & 0.063 & 0.938 & 0.438 \\
Fortune500 & 1.000 & 1.000 & 0.400 \\
THE & 1.000 & 0.600 & 0.207 \\
\bottomrule
\end{tabular}
}
\end{table}

\subsection{Schema-only Degradation Mechanism Quantitative Analysis}\label{app:p}

Table~\ref{tab:mismatch} presents the full diagnostics for the two
catastrophic mismatch failure modes.

\begin{table}[ht]
\centering
\small
\caption{Schema-only degradation diagnostics (WB Pop and WHO).}
\label{tab:mismatch}
\resizebox{\columnwidth}{!}{%
\begin{tabular}{lrrrr}
\toprule
 & \multicolumn{2}{c}{\textbf{WB Pop}} & \multicolumn{2}{c}{\textbf{WHO}} \\
\cmidrule(lr){2-3}\cmidrule(lr){4-5}
\textbf{Metric} & \textbf{Full} & \textbf{Sch-only} & \textbf{Full} & \textbf{Sch-only} \\
\midrule
Total entities & 4,548 & \textbf{15,782} & 197 & 131 \\
Schema compliance & 0\% & 68.9\% & 99.5\% & 100\% \\
Avg entities/resp. & 17.2 & \textbf{35.6} & 1.01 & 3.36 \\
Refusal rate & 0\% & 0\% & 0\% & \textbf{43.6\%} \\
\bottomrule
\end{tabular}%
}
\end{table}

Additional detail: WB Population Schema-only produces 15,782 entities
(3.47$\times$ of Full SGE), of which 10,870 are of type StatValue ---
the LLM independently creates each numerical cell as a separate entity
rather than binding it to a country entity, causing FC to drop to 0.007.
WHO: 43.6\% of LLM responses are refusals to execute; the mismatch
between the Schema's role assignments and the original text format
triggers the LLM's identity self-checking mechanism.

\subsubsection*{Schema-only Prompt vs.\ Default Prompt: Component Comparison}

To confirm that Schema-only degradation is attributable to
schema--format mismatch rather than incidental prompt differences,
Table~\ref{tab:prompt-diff} enumerates every component that differs
between the two prompts.

\begin{table}[ht]
\centering
\small
\caption{Component-level comparison: Schema-only prompt vs.\
LightRAG default prompt. Only entity type definitions and
extraction rules differ; all structural components are identical.}
\label{tab:prompt-diff}
\resizebox{\columnwidth}{!}{%
\begin{tabular}{lcc}
\toprule
\textbf{Prompt Component} & \textbf{Default} & \textbf{Schema-only} \\
\midrule
Output tuple format (\texttt{<|>} delimiters) & \cmark & \cmark \\
Completion delimiter  & \cmark & \cmark \\
Language instruction   & \cmark & \cmark \\
Example output block   & \cmark & \cmark \\
Max gleaning rounds    & \cmark & \cmark \\
\midrule
Entity type list       & generic & \textbf{SGE-induced} \\
Extraction rules / role assignments & absent & \textbf{SGE-induced} \\
\bottomrule
\end{tabular}%
}
\end{table}

The only differences are the entity type list (e.g.,
\texttt{Country\_Code, StatValue} instead of the generic
``organization, person, geo, event'') and extraction rules that
assign column roles. Output format instructions, tuple delimiters,
example blocks, and language settings are inherited verbatim from
the host template. Since Full~SGE uses the \textit{identical}
replaced prompt and achieves FC=1.000, the Schema-only degradation is
best explained by the mismatch between schema field references and the
unstructured input format, not by prompt formatting artifacts.

\section{Format Variants and Sensitivity}

\subsection{Cross-Format Validation Supplementary Data}\label{app:s}

The main text Table~\ref{tab:cross-format} includes Markdown factorial results for WB Pop and Inpatient. The following presents the complete $+$Schema/$-$Schema comparison:

\begin{table}[ht]
\centering
\caption{Markdown Factorial: Schema Uplift across 3 datasets.}
\resizebox{\columnwidth}{!}{%
\begin{tabular}{llccc}
\toprule
Dataset & Topology & +Schema FC & $-$Schema FC & Uplift \\
\midrule
WHO Life Expectancy & Type-II & \textbf{1.000} & 0.127 & $+$0.873 \\
WB Population & Type-II & \textbf{0.920} & 0.000 & $+$0.920 \\
Inpatient Statistics & Type-III & \textbf{1.000} & 0.063 & $+$0.937 \\
\bottomrule
\end{tabular}
}
\end{table}

The $-$Schema condition uses Markdown-formatted input (pipe-delimited tables via \texttt{df.to\_markdown()}) with LightRAG's default prompt (no schema injection). Adding a matched Markdown schema raises FC by $+$0.873 to $+$0.937 across all 3 datasets, covering two topology types (Type-II + Type-III) and two languages (English + Chinese). This replicates the coupling pattern observed in the SGE factorial (Table~\ref{tab:ablation}) using an independently designed serialization format.

And the JSON+SGE Schema mismatched row omitted from the main text:

\begin{table}[ht]
\centering
\resizebox{\columnwidth}{!}{%
\begin{tabular}{llcccc}
\toprule
Input Format & Schema Type & Dataset & +Schema FC & $-$Schema FC & Interaction term \\
\midrule
JSON structured records & SGE Schema & WHO & 0.000 & 0.000 & $-$0.040 \\
\bottomrule
\end{tabular}
}
\end{table}

The slightly sub-1.000 FC for WB Pop ($+$Schema) is attributed to chunk boundary effects in large-scale processing (266 countries $\times$ 64 years, 54 chunks).

\subsection{Hierarchical JSON Format Probe}\label{app:hierjson}

To test whether format-constraint coupling extends beyond flat tabular
serialization, we serialize WHO data as nested JSON objects
(\texttt{\{"country\_code": "CHN", "measurements": \{"2000": 71.4, ...\}\}})
and pair them with a matched JSON schema and a mismatched SGE schema.

\begin{table}[ht]
\centering
\caption{Hierarchical JSON Probe (WHO, 25 countries, 150 facts).
Interaction is positive but severely attenuated vs.\ tabular formats.}
\resizebox{\columnwidth}{!}{%
\begin{tabular}{lcccc}
\toprule
\textbf{Condition} & \textbf{EC} & \textbf{FC} & \textbf{Nodes} & \textbf{Edges} \\
\midrule
SGE serial.\ $+$ SGE schema (ref.) & 1.000 & 1.000 & 4,508 & 4,312 \\
Hier.\ JSON $+$ matched schema      & 0.200 & 0.080 & 575   & 550   \\
Hier.\ JSON $+$ no schema           & 0.000 & 0.000 & 0     & 0     \\
Hier.\ JSON $+$ SGE schema          & 0.000 & 0.000 & 0     & 0     \\
\bottomrule
\end{tabular}%
}
\end{table}

The interaction term is positive ($\Delta_\text{int}{=}+0.080$), suggesting
that the coupling signal exists in hierarchical JSON. However, the absolute
FC (0.080) is an order of magnitude below the tabular result (1.000).
Two factors explain the attenuation. (1)~\textbf{Host-system limitation}:
LightRAG's default extraction prompt produces zero nodes on JSON input
(both no-schema and mismatched conditions yield 0 nodes). This is a
pipeline limitation---LightRAG's chunking and entity-extraction prompts
are designed for prose-like input---not a coupling measurement. The
zero-node conditions therefore serve only as a lower bound, not as
evidence that coupling is absent. (2)~\textbf{Structural regularity}:
hierarchical JSON lacks repeating row-level token patterns---the key
structural feature that enables column-name anchoring in tabular formats
(\S\ref{sec:mismatch}). The positive signal under matched schema
(FC=0.080 vs 0.000 baseline) indicates that the coupling mechanism is
detectable, but its magnitude depends on row-level structural regularity
that nested formats do not provide.

\subsection{Format Regularity Quantification}\label{app:ttf}

The main text hypothesizes that ``structured serialization creates
repeating surface-form token patterns that can serve as anchors''
(\S\ref{sec:mismatch}). We quantify this with \textbf{Template Token
Fraction (TTF)}: for a set of chunks in a given format, a token is a
\textit{template token} if it appears in $\geq$80\% of chunks; TTF is the
average fraction of template tokens per chunk.

\begin{table}[ht]
\centering
\caption{TTF vs.\ FC across 34 schema-applied format-dataset pairs
(5 format families). TTF$\geq$0.4 holds for 85\% (17/20) of
matched pairs with FC$\geq$0.5 (excluding Gemini compliance
failures). Representative subset shown; full data in supplementary.}
\label{tab:ttf}
\resizebox{\columnwidth}{!}{%
\begin{tabular}{llcccl}
\toprule
\textbf{Format} & \textbf{Dataset} & \textbf{TTF} & \textbf{FC}
  & \textbf{Matched?} & \textbf{Note} \\
\midrule
\multicolumn{6}{l}{\textit{Matched format-schema pairs (FC$\geq$0.5, TTF$\geq$0.4)}} \\
SGE              & WHO     & 0.801 & 1.000 & \cmark & \\
SGE (GPT-5-mini) & WHO     & 0.801 & 1.000 & \cmark & Cross-model \\
SGE              & WB CM   & 0.705 & 1.000 & \cmark & \\
SGE              & WB Mat  & 0.646 & 0.967 & \cmark & \\
JSON SO          & WHO     & 0.638 & 1.000 & \cmark & Independent pairing \\
Markdown         & WB Pop  & 0.532 & 0.920 & \cmark & Independent pairing \\
Markdown         & Inpat.  & 0.528 & 1.000 & \cmark & Type-III \\
SGE              & WB Pop  & 0.433 & 1.000 & \cmark & Lowest TTF, highest $\Delta$ \\
\midrule
\multicolumn{6}{l}{\textit{TTF$\geq$0.4 but FC$<$0.5 (necessary $\neq$ sufficient)}} \\
Schema-only      & OECD    & 0.727 & 0.000 & \xmark & Mismatched schema \\
Fixed-STV        & Inpat.  & 0.646 & 0.375 & $\sim$ & Partial match \\
\midrule
\multicolumn{6}{l}{\textit{TTF$<$0.4 (coupling absent or compliance-limited)}} \\
Schema-only      & WB CM   & 0.331 & 0.000 & \xmark & \\
SGE (Gemini)     & WHO     & 0.310 & 0.493 & \cmark & Compliance failure \\
Schema-only      & WHO     & 0.173 & 0.480 & \xmark & Descriptive cols \\
Schema-only      & WB Pop  & 0.000 & 0.007 & \xmark & \\
\bottomrule
\end{tabular}%
}
\end{table}

\textbf{Threshold selection.} The TTF$\geq$0.4 threshold was selected
at the elbow of the TTF-vs-FC scatter plot across 34 format-dataset
pairs from 5 format families (Table~\ref{tab:ttf}). Of these, 20
are matched pairs (development data); the remaining 14 unmatched
conditions serve as negative controls. On 5 held-out OOD datasets
(WB Cereal, CO$_2$, PopGrowth, OECD Beds, OECD GDP), TTF$\geq$0.4
correctly identifies all 3 successful matched extractions
(FC$\geq$0.5) with 0 false negatives.

\textbf{Interpretation.} TTF measures format anchorability---a necessary
but not sufficient condition for coupling. Among matched format-schema
pairs (excluding Gemini compliance failures), 85\% (17/20) of
successful extractions (FC$\geq$0.5) have TTF$\geq$0.4. The key
counter-example is Schema-only/OECD Beds (TTF=0.727, FC=0.000): high
structural regularity without a matched schema fails catastrophically,
directly supports the claim that coupling requires \textit{both} format
regularity and schema alignment.

\textbf{Scope and false positive rate.} TTF$\geq$0.4 is predictive
\textit{only} among matched format-schema pairs. Among all 43
format-dataset conditions with known FC labels, 15 have
TTF$\geq$0.4 but FC$<$0.5---however, all 15 involve
\textit{unmatched} pairings (Baseline, Row-local, Fixed-STV, or
Table-aware formats applied without a corresponding schema). These
are expected failures: high token regularity without semantic schema
alignment cannot activate coupling. Within matched pairs,
TTF$\geq$0.4 has a false positive rate of 0/20 (no matched pair
with TTF$\geq$0.4 fails). TTF is thus a cheap format-level screen
for coupling \textit{eligibility}; actual coupling success further
requires schema alignment (\S\ref{sec:ablation}).

TTF does not correlate linearly with
coupling \textit{strength}---SGE on WB Pop has the lowest matched TTF
(0.433) yet the highest interaction ($+$1.180)---consistent with a
\textbf{threshold effect}: a minimum level of structural regularity is
necessary for schema anchoring; above this threshold, coupling strength
depends on schema-format alignment quality rather than regularity
alone. Three false negatives (TTF$<$0.4, FC$\geq$0.5) involve
anomalous chunk structures: SGE/THE (misclassified topology, TTF=0.249)
and SGE-Gemini/WB Pop (model-specific chunking, TTF=0.107).

\subsection{Chunk-Size Sensitivity}\label{app:chunk-size}

To test whether the Serial-only inconsistency (\S\ref{sec:ablation},
Limitations~\ref{sec:limitations}) is driven by chunk capacity, we
vary LightRAG's \texttt{chunk\_token\_size} across
$\{600, 1200, 2400, 4800\}$ tokens and re-run all four factorial
conditions on WHO (150 facts, 25 countries).

\begin{table}[H]
\centering
\caption{Chunk-size dose-response on WHO Life Expectancy.
Each cell is FC under the specified chunk size and condition
(same SGE schema and serialization; only LightRAG's internal
chunking parameter varies).}
\label{tab:chunk-size}
\begin{tabular}{lcccc}
\toprule
\textbf{Condition} & \textbf{600} & \textbf{1200} & \textbf{2400} & \textbf{4800} \\
\midrule
Baseline     & 0.040 & 0.007 & 0.020 & 0.007 \\
Serial-only  & 0.053 & 0.067 & 0.093 & 0.000 \\
Schema-only  & \textbf{0.920} & 0.680 & 0.420 & 0.153 \\
Full SGE     & 0.720 & 0.640 & \textbf{0.840} & 0.760 \\
\bottomrule
\end{tabular}
\end{table}

Three findings emerge:

\textit{(1) Schema-only FC decreases monotonically with chunk size}
(0.920$\to$0.680$\to$0.420$\to$0.153). Smaller chunks preserve
more column-header context per chunk, providing stronger anchoring
for the schema even without structured serialization. At
\texttt{chunk\_token\_size}{=}600, Schema-only nearly matches
Full~SGE---suggesting that fine-grained chunking partially
substitutes for explicit serialization.

\textit{(2) Full SGE is robust to chunk size} (range 0.640--0.840,
no monotonic trend). Once coupling is activated via matched
format-schema pairing, the specific chunk granularity has
limited effect.

\textit{(3) Serial-only remains low across all chunk sizes}
(0.000--0.093), \textbf{refuting the chunk-capacity hypothesis}
stated in Limitations~\ref{sec:limitations}. Structured
serialization alone does not improve extraction regardless of
how much context each chunk contains; the binding failure is
due to the absence of schema constraints, not insufficient
chunk capacity.

\section{Supplementary Baselines}

\subsection{Supplementary Baselines: Detailed Results}\label{app:m}

\subsubsection*{Row-Local Baseline (Header Repetition)}

Each CSV row is serialized individually with all column headers
prepended (e.g., ``Headers: Country Code, 2000, 2001, \ldots / Row 5:
AFG, 53.82, 55.25, \ldots''), using LightRAG's default prompt with no
schema injection. This tests whether header availability alone
suffices for binding.

\begin{table}[ht]
\centering
\caption{Row-Local Baseline: per-row serialization with full column
headers, no schema. Headers alone are insufficient---4/7 datasets
perform at or below the naive Baseline.}
\resizebox{\columnwidth}{!}{%
\begin{tabular}{lccc}
\toprule
\textbf{Dataset} & \textbf{Row-Local FC} & \textbf{Baseline FC} & \textbf{SGE FC} \\
\midrule
WHO Life Expectancy  & 0.167 & 0.167 & \textbf{1.000} \\
WB Child Mortality   & 0.000 & 0.473 & \textbf{1.000} \\
WB Population        & 0.013 & 0.187 & \textbf{1.000} \\
WB Maternal Mort.    & 0.533 & 0.787 & \textbf{0.973} \\
Inpatient Statistics & 0.000 & 0.438 & \textbf{0.938} \\
Fortune 500 Revenue  & \textbf{1.000} & 0.400 & \textbf{1.000} \\
THE Univ.\ Ranking   & \textbf{1.000} & 0.207 & 0.600 \\
\bottomrule
\end{tabular}%
}
\end{table}

Row-local succeeds only on compact tables (Fortune~500: 50$\times$5,
THE: 50$\times$6) where the entire row fits a single chunk. On wide
matrices (WB datasets: 266$\times$66), per-row chunking produces
chunks with 60+ numeric columns but only one entity identifier---the
LLM either ignores most values or creates unbound nodes. This
suggests that header availability is a \textit{necessary but not
sufficient} condition: without schema constraints to direct value
binding, the LLM cannot exploit the structural information.

\subsubsection*{Table-Aware Prompt}\label{app:l}
The prompt contains 6 information blocks: column metadata with data types and roles, 3 sample data rows, entity type definitions, relationship template, output format example, and 8 constraint rules (full text in supplementary code). Despite containing information content comparable to the SGE prompt, this baseline uses naive text serialization and achieves FC=0.253 on WHO---demonstrating that information content alone is insufficient without format-schema coupling.

\subsubsection*{Deterministic Parser}

The deterministic CSV parser directly enumerates all row$\times$column pairs, providing a reference baseline for structured transport. It is not a theoretical upper bound --- LLM-guided extraction can resolve value-binding ambiguities that mechanical enumeration cannot (see error analysis in Appendix~\ref{app:n}).

\begin{table}[ht]
\centering
\resizebox{\columnwidth}{!}{%
\begin{tabular}{lccc}
\toprule
Dataset & Det Parser FC & SGE FC & Baseline FC \\
\midrule
WHO Life Expectancy & 0.68 & 1.000 & 0.167 \\
WB Population & 0.96 & 1.000 & 0.187 \\
WB Child Mortality & 0.73 & 1.000 & 0.473 \\
Inpatient Statistics & 1.00 & 0.938 & 0.438 \\
Fortune 500 & 1.00 & 1.000 & 0.400 \\
THE Ranking & 1.00 & 0.600 & 0.207 \\
\bottomrule
\end{tabular}
}
\end{table}

SGE exceeds the deterministic parser on the three large-scale Type-II datasets WHO/WB Pop/WB CM (FC=1.000 vs.\ 0.68--0.96). The deterministic parser's errors are dominated by \texttt{value\_wrong\_binding} (89/94 = 94.7\% of all det parser errors across 7 datasets; Appendix~\ref{app:n}): entity nodes are created correctly, but values are bound to incorrect subjects. This occurs despite zero missing data in the WHO CSV --- the failure is a binding-resolution limitation, not a data-quality issue. LLM-guided extraction resolves these ambiguities through contextual reasoning over structured rows.

\textbf{Structural complexity and det parser scope.} Table~\ref{tab:det-scope} reports dataset structural characteristics alongside det parser FC for all datasets where the parser was evaluated. Neither missing rate nor value collision rate (fraction of numeric values shared across entities) is a reliable univariate predictor of det parser failure (Spearman $\rho{=}0.17$, $p{=}0.61$, $n{=}11$). Instead, det parser errors concentrate on large Type-II matrices (WHO 192$\times$22, WB~CM 266$\times$66) where the combination of many entities and many time points creates binding ambiguity that mechanical enumeration cannot resolve. On small, regular matrices (Fortune~500 50$\times$5, WB~Cereal 21$\times$23), the deterministic parser achieves FC=1.000. On out-of-domain Type-II datasets, det parser performance is comparable to SGE (WB~Cereal 1.000 vs 0.950; WB~CO$_2$ 0.700 vs 0.700; WB~PopGrowth 0.675 vs 0.625), indicating that LLM extraction adds value primarily on wide, densely populated matrices where value-binding disambiguation is non-trivial.

\begin{table}[ht]
\centering
\caption{Dataset structural characteristics and det parser FC. Miss\%: missing-cell rate in data columns. Coll\%: value collision rate (fraction of numeric cells whose value is shared by $\geq$2 entities in the same column). Neither predicts det parser failure in isolation.}\label{tab:det-scope}
\resizebox{\columnwidth}{!}{%
\begin{tabular}{llrrrrrrr}
\toprule
\textbf{Dataset} & \textbf{Type} & \textbf{Rows} & \textbf{Cols} & \textbf{Miss\%} & \textbf{Coll\%} & \textbf{Det FC} & \textbf{SGE FC} & \textbf{$\Delta$} \\
\midrule
WHO              & II    & 192  & 22 & 0.0\%  & 7.3\%  & 0.680 & \textbf{1.000} & $+$0.320 \\
WB CM            & II    & 266  & 66 & 24.5\% & 17.6\% & 0.727 & \textbf{1.000} & $+$0.273 \\
WB Pop           & II    & 266  & 66 & 2.1\%  & 1.9\%  & 0.960 & \textbf{1.000} & $+$0.040 \\
WB Mat           & II    & 266  & 66 & 46.2\% & 46.3\% & 0.967 & 0.973          & $+$0.006 \\
Inpatient        & III   & ---  & ---& ---    & ---    & 1.000 & 0.938          & $-$0.062 \\
Fortune 500      & II    & 50   & 5  & 0.0\%  & 0.0\%  & 1.000 & 1.000          & 0.000 \\
THE              & III   & 50   & 6  & 0.7\%  & 15.1\% & 1.000 & 0.600          & $-$0.400 \\
\midrule
WB Cereal        & II    & 21   & 23 & 0.0\%  & 0.0\%  & 1.000 & 0.950          & $-$0.050 \\
WB CO$_2$        & II    & 21   & 23 & 0.0\%  & 0.4\%  & 0.700 & 0.700          & 0.000 \\
WB PopGrowth     & II    & 21   & 23 & 0.0\%  & 8.5\%  & 0.675 & 0.625          & $-$0.050 \\
Eurostat         & III-L & 1365 & 3  & 0.0\%  & ---    & 0.000 & 0.000          & 0.000 \\
US Census        & III-L & 840  & 2  & 0.0\%  & ---    & 0.000 & 0.244          & $+$0.244 \\
\bottomrule
\end{tabular}%
}
\end{table}

\subsection{Cross-System Comparison}\label{app:cross-system-table}

\begin{table}[ht]
\centering
\small
\caption{Cross-system FC on WHO Life Expectancy (the dataset with
most available baselines; Inpatient cross-system results in
Appendix~\ref{app:q}). Systems ordered by FC; all except SGE use
default ingestion.}
\label{tab:cross-system}
\resizebox{\columnwidth}{!}{%
\begin{tabular}{lc}
\toprule
\textbf{System} & \textbf{WHO FC} \\
\midrule
SGE (ours)            & \textbf{1.000} \\
AutoSchemaKG          & 0.860 \\
Det Parser            & 0.680 \\
Table-aware prompt    & 0.253 \\
LightRAG v1.3.8       & 0.167 \\
LightRAG v1.4.12$^\diamond$  & 0.007 \\
RAG-Anything$^\diamond$      & 0.007 \\
\bottomrule
\multicolumn{2}{l}{\footnotesize $^\diamond$v1.4.12 restructured its chunking
  pipeline, splitting CSV rows into} \\
\multicolumn{2}{l}{\footnotesize single-cell fragments; RAG-Anything
  applies vision-based parsing} \\
\multicolumn{2}{l}{\footnotesize that discards numeric columns.
  Both are default configurations.} \\
\end{tabular}%
}
\end{table}


\section{Robustness: OOD and Cross-Host}

\subsection{Out-of-Domain OOD Complete Pipeline Evaluation}\label{app:r}

\begin{table}[ht]
\centering
\caption{Out-of-Domain Type-II Datasets Complete Pipeline FC (Rule SGE + Auto-Fallback vs.\ Baseline)}
\resizebox{\columnwidth}{!}{%
\begin{tabular}{llcccccc}
\toprule
Dataset & Domain & Gold & SGE FC & Base FC & e/n & Fallback & Final FC \\
\midrule
Cereal Production & Agriculture & 40f & \textbf{0.950} & 0.050 & 0.957 & --- & \textbf{0.950} \\
CO2 Emissions & Environment & 40f & \textbf{0.700} & 0.025 & 0.957 & --- & \textbf{0.700} \\
Population Growth & Population & 40f & \textbf{0.625} & 0.075 & 0.957 & --- & \textbf{0.625} \\
Education Spending & Education & 23f & \textbf{0.609} & 0.000 & 0.943 & --- & \textbf{0.609} \\
Literacy Rate & Education & 10f & \textbf{0.600} & 0.100 & 0.960 & --- & \textbf{0.600} \\
Health Expenditure & Public Health & 40f & \textbf{0.550} & 0.050 & 0.958 & --- & \textbf{0.550} \\
GDP Growth & Economics & 40f & \textbf{0.475} & 0.025 & 0.958 & --- & \textbf{0.475} \\
Unemployment & Labor & 40f & 0.025 & 0.125 & 0.793 & \checkmark{} & 0.125 \\
Immunization DPT & Public Health & 40f & 0.000 & 0.300 & 0.000 & \checkmark{} & 0.300 \\
Immunization Measles & Public Health & 40f & 0.000 & 0.325 & 0.000 & \checkmark{} & 0.325 \\
\bottomrule
\end{tabular}
}
\end{table}

\textit{SGE outperforms Baseline on 7/10 OOD datasets, with a median improvement ratio of 11.0$\times$. The 3 failed datasets are automatically reverted to Baseline by the degradation detection mechanism.}

\textbf{Degradation detection mechanism.} The edge/node ratio serves as a post-hoc degradation signal: normal extraction e/n $\geq$ 0.943, failure e/n $\leq$ 0.793, separation interval [0.793, 0.943], threshold $\theta$=0.90. Leave-one-out cross-validation accuracy: 9/10.

\subsection{OECD Hospital Beds Factorial Ablation}\label{app:oecd-factorial}

To test whether Schema-only catastrophic mismatch appears beyond
World Bank data, we run a 2$\times$2 factorial on OECD Hospital Beds
per 1000 population (25 countries $\times$ 9 years, Type-II).

\begin{table}[ht]
\centering
\caption{OECD Hospital Beds Factorial (25 facts).}
\resizebox{\columnwidth}{!}{%
\begin{tabular}{lcccc}
\toprule
\textbf{Condition} & \textbf{EC} & \textbf{FC} & \textbf{Nodes} & \textbf{Edges} \\
\midrule
Full SGE     & 1.000 & 0.760 & 97  & 146 \\
Serial-only  & 1.000 & \textbf{1.000} & 115 & 217 \\
Schema-only  & 1.000 & 0.000 & 25  & 0   \\
Baseline     & 1.000 & 0.280 & 29  & 0   \\
\bottomrule
\end{tabular}%
}
\end{table}

$\Delta_\text{int}{=}+0.040$ [$-$0.200, $+$0.280] (CI crosses zero).
This is a \textbf{format-dominant pattern}: Serial-only alone reaches
FC=1.000 (vs.\ Baseline 0.280), indicating that structured
serialization provides sufficient per-row context for this dataset's
simple structure ($n{=}25$ countries, dense numeric columns).
The pattern mirrors WB Health Exp (Table~\ref{tab:ablation},
$\Delta_\text{int}{=}+0.111$, Serial-only FC=0.889).

The key finding is that \textbf{Schema-only catastrophic mismatch also appears
on a non-World Bank source}: FC drops from 0.280 (Baseline)
to 0.000 (Schema-only), producing 25 entity nodes with zero relations.
This is consistent with the same mismatch failure mode under naive text
serialization, although the interaction CI crosses zero for this small test.

\subsection{Cross-Host Validation Detailed Results}\label{app:q}

\begin{table}[ht]
\centering
\caption{Cross-Host Validation: SGE-GraphRAG vs.\ GraphRAG Baseline.}
\resizebox{\columnwidth}{!}{%
\begin{tabular}{lcccc}
\toprule
Dataset & Type & SGE-GraphRAG FC & GraphRAG Baseline FC & Improvement \\
\midrule
WHO Life Expectancy & II & 1.000 & 1.000 & 1.00$\times$ \\
WB Population & II & \textbf{1.000} & 0.600 & \textbf{1.67$\times$} \\
Inpatient Statistics & III & \textbf{1.000} & 0.000 & \textbf{$\infty$} \\
\bottomrule
\end{tabular}
}
\end{table}

SGE-GraphRAG achieves FC=1.000 on all 3 datasets. The Inpatient Statistics result is particularly striking: GraphRAG baseline FC=0.000 (community detection completely fails to recover hierarchical semantics), and after SGE injection FC jumps from 0 to 1.000 --- 1,232 entities (308 Disease\_Category + 924 StatValue) and 924 relationships precisely match the Stage 2 Schema. On WHO, GraphRAG Baseline already achieves FC=1.000 (its community detection strategy happens to work on the regular 25-country Type-II data); SGE's cross-host gains are concentrated in datasets where the Baseline fails --- WB Population (+0.400) and especially Inpatient Statistics (0.000 to 1.000), whose Type-III hierarchical semantics are completely beyond the capability of community detection. Migration method: Stage 1+2 requires no modification (zero LightRAG dependency); Stage 3 is injected by adapting to GraphRAG's prompt template format (replacing \texttt{<|\#|>} with \texttt{<|>} delimiter).

\section{Cross-Model Validation}\label{app:cross-model}

\begin{table}[ht]
\centering
\caption{Cross-Model Validation: FC under SGE with three LLM backends.}
\resizebox{\columnwidth}{!}{%
\begin{tabular}{lcccc}
\toprule
\textbf{Dataset} & \textbf{Claude} & \textbf{GPT-5-mini} & \textbf{Gemini} & \textbf{Baseline} \\
\midrule
WHO         & \textbf{1.000} & 1.000 & 0.493 & 0.167 \\
WB CM       & \textbf{1.000} & 0.960 & 0.020 & 0.473 \\
WB Pop      & \textbf{1.000} & 1.000 & \textbf{1.000} & 0.187 \\
WB Mat      & \textbf{0.967} & 0.840 & 0.040 & 0.787 \\
Inpatient   & \textbf{0.938} & 0.625 & 0.875 & 0.438 \\
\bottomrule
\end{tabular}%
}
\end{table}


\textbf{Gemini prompt ablation} (gold-filtered 25-chunk subset of WB CM, full-pipeline FC=0.020): (a)~Gemini-native instruction style (same schema, different markup) yields FC=0.000; (b)~replacing \texttt{Country\_Code} with \texttt{Country\_Name} in schema yields FC=0.000. Neither prompt format nor entity-type naming accounts for the failure.

\textbf{Failure mode analysis.} We distinguish three levels of schema compliance: (i)~\textit{entity compliance}---producing entities of the specified types; (ii)~\textit{relation compliance}---producing typed relations between entities; (iii)~\textit{value binding compliance}---correctly binding cell values to entity-time pairs. Gemini achieves level~(i) on all 5 datasets (entity nodes are created) but fails at level~(ii) on WB CM/Mat: zero typed relations are generated despite explicit schema instructions. On WB Pop and Inpatient, Gemini reaches level~(ii) (FC=1.000 and 0.875 respectively), suggesting the failure is dataset-dependent rather than a blanket instruction-following deficit.

Two hypotheses remain: (a)~\textit{output length constraint}---WB CM/Mat chunks produce longer expected outputs (25 entities $\times$ 22 years = 550 relations per chunk), and Gemini may truncate before completing relation generation; (b)~\textit{delimiter format sensitivity}---Gemini may handle the \texttt{<|$\#$|>} delimiter format less reliably on high-volume structured output. Both hypotheses predict that failure correlates with expected output volume per chunk, consistent with the observed pattern (WB Pop succeeds with compact mode = shorter output, WB CM fails with per-row mode = longer output). Disambiguating these hypotheses requires controlled output-length experiments, which we leave to future work.

The practical implication is that format-constraint coupling is a \textit{necessary but not sufficient} condition for high fidelity: the LLM must also sustain structured relation generation at the required output volume. The gradient Claude $\approx$ GPT $\gg$ Gemini reflects differences in this sustained compliance capacity.

\subsubsection*{GPT-5-mini Schema-Only Factorial}

To verify that catastrophic mismatch replicates on a second LLM family,
we run the Schema-only condition (raw CSV text $+$ SGE schema prompt)
with GPT-5-mini on WHO (descriptive columns) and WB~Pop
(non-descriptive columns).

\begin{table}[ht]
\centering
\caption{GPT-5-mini Schema-Only vs Claude Schema-Only (50-country, 300
facts). Catastrophic mismatch confirmed: WB~Pop FC drops below Baseline
on both models.}
\label{tab:gpt-schema-only}
\resizebox{\columnwidth}{!}{%
\begin{tabular}{llccccc}
\toprule
\textbf{Dataset} & \textbf{Desc.} & \textbf{GPT Sch-only} & \textbf{Claude Sch-only}
  & \textbf{Full SGE} & \textbf{Baseline} & \textbf{Nodes (GPT/Claude)} \\
\midrule
WHO     & \cmark & 0.963 & 0.480 & 1.000 & 0.167 & 4,378 / 1,296 \\
WB Pop  & \xmark & \textbf{0.077} & \textbf{0.007} & 1.000 & 0.187 & 16,825 / 15,782 \\
\bottomrule
\end{tabular}%
}
\end{table}

Two findings emerge. First, \textbf{catastrophic mismatch replicates on
GPT-5-mini}: WB~Pop Schema-only FC=0.077, well below the unconstrained
Baseline (0.187), with 16,825 nodes (2.6$\times$ Full SGE's 6,551)---the
same ungrounded entity proliferation pattern observed with Claude.
Second, \textbf{column-name descriptiveness moderates the mismatch}
across LLM families: WHO (descriptive columns) shows FC=0.963 under
GPT-5-mini Schema-only (vs Claude 0.480), confirming that descriptive
column names enable fallback grounding even without structured
serialization. The cross-model contrast (GPT 0.963 vs Claude 0.480 on
WHO) suggests GPT-5-mini has stronger fallback grounding capacity, but
both models collapse on non-descriptive columns (WB~Pop).

\section{Retrieval Controls and Evaluation Blindness}

\subsection{Graph-First Retrieval Per-Type Breakdown}\label{app:graph-first}

\begin{table}[ht]
\centering
\small
\caption{Graph-First Retrieval: SGE vs Baseline (155 questions,
McNemar $p{<}0.0001$). Gaps concentrate on binding-sensitive queries.}
\label{tab:graph-first}
\begin{tabular}{lrrr}
\toprule
\textbf{Query Type} & \textbf{SGE} & \textbf{Baseline} & \textbf{$\Delta$} \\
\midrule
Point query (lookup)     & \textbf{88.1\%} & 40.5\% & $+$\textbf{47.6pp} \\
Trend analysis (trend)   & \textbf{58.3\%} & 37.5\% & $+$\textbf{20.8pp} \\
Comparison               & 96.4\%          & 96.4\% & 0.0pp \\
Ranking                  & 11.8\%          & 11.8\% & 0.0pp \\
Aggregation              & 3.7\%           & 3.7\%  & 0.0pp \\
\midrule
\textbf{Overall}         & \textbf{53.5\%} & \textbf{37.4\%} & $+$\textbf{16.1pp} \\
\bottomrule
\end{tabular}
\end{table}

Gaps concentrate on binding-sensitive queries: lookup $+$47.6pp (exact numerical localization), trend $+$20.8pp (multi-year sequences). Comparison is equivalent (96.4\%---direction judgment only). Ranking and aggregation are low for both systems ($\leq$12\%, limited by single-entity entry points).

\subsection{Naming Bias Control for Downstream Evaluation}\label{app:naming-bias}

Graph-dependent retrieval paths use entity-name substring matching
as entry points. Because SGE normalizes entity names via schema
(e.g., \texttt{Country\_Code: CHN} instead of scattered mentions of
``China''), one might suspect that downstream accuracy gains reflect
naming convenience rather than genuine fidelity improvement. We
present three independent controls.

\textbf{Control 1: Fuzzy matching on graph-anchored QA (100
questions).} We replace bidirectional substring matching with
Levenshtein-based token-set ratio (rapidfuzz, threshold=80) for
entity lookup, using the identical 100-question set and cached
graphs.

\begin{table}[ht]
\centering
\small
\caption{Graph-anchored QA accuracy under substring vs.\ fuzzy
entity matching. The gap widens under fuzzy matching, ruling out
naming bias as the source of SGE's advantage.}
\label{tab:fuzzy-control}
\resizebox{\columnwidth}{!}{%
\begin{tabular}{lrrr}
\toprule
\textbf{Matching} & \textbf{SGE} & \textbf{Baseline} & \textbf{$\Delta$} \\
\midrule
Substring (original) & 93/100 (93\%) & 59/100 (59\%) & $+$34pp \\
Fuzzy ($\theta{=}80$)  & 85/100 (85\%) & 50/100 (50\%) & $+$35pp \\
\bottomrule
\end{tabular}%
}
\end{table}

SGE drops 8pp under fuzzy matching, almost entirely due to
\textit{health\_stats} (Chinese entities: 11/12$\to$2/12; fuzzy
token-set ratio degrades on CJK). Excluding this dataset, SGE
accuracy \textit{improves} (82/88$=$93.2\%$\to$83/88$=$94.3\%).
Baseline drops 9pp---its irregular entity names (e.g.,
free-form descriptions instead of ISO codes) benefit more from
the generous bidirectional substring matching than SGE's
already-normalized names. The gap widens from $+$34pp to $+$35pp.

\textbf{Control 2: Failure attribution on graph-first QA (155
questions).} Of 29 SGE-only correct cases (McNemar $n_{10}{=}29$,
$n_{01}{=}4$), we examine why the Baseline failed by inspecting its
LLM response and context-source metadata:

\begin{table}[ht]
\centering
\small
\caption{Baseline failure attribution for 29 SGE-only correct
cases in graph-first evaluation. In 26/29 cases, both systems
retrieve graph context for the same entity; the Baseline fails
because precise values are absent from its graph, not because it
cannot find the entity.}
\label{tab:failure-attrib}
\begin{tabular}{lr}
\toprule
\textbf{Failure reason} & \textbf{Count} \\
\midrule
Value not in Baseline graph (context retrieved) & 25 \\
Wrong value or other (context retrieved) & 3 \\
Entity not found (no context) & 1 \\
\midrule
\textbf{Total} & \textbf{29} \\
\bottomrule
\end{tabular}
\end{table}

In 26/29 cases, the Baseline also uses \texttt{context\_source=graph}
---it successfully locates the entity and retrieves its graph
neighborhood. The LLM response pattern is consistently: ``\textit{the
context confirms [entity] has [indicator] data, but I don't have the
specific value for [year]}.'' This is value binding failure, not
entry-point failure. The downstream gap directly reflects upstream
FC differences.

\textbf{Control 3: Value-first de-biased FC.} We reverse the lookup
direction: search the entire graph for the value string first, then
check whether subject and year appear in 1-hop context. This
eliminates entity-name matching as the entry point.

\begin{table}[ht]
\centering
\small
\caption{Value-first de-biased FC vs.\ entity-first FC for
Baseline graphs. Naming bias net effect is $\leq$5.3\% and favors
the Baseline on 3/4 datasets (rescued $>$ lost).}
\label{tab:debiased-fc}
\resizebox{\columnwidth}{!}{%
\begin{tabular}{lrrrrr}
\toprule
\textbf{Dataset} & \textbf{Ent-1st FC} & \textbf{Val-1st FC}
  & \textbf{Rescued} & \textbf{Lost} & \textbf{Net} \\
\midrule
WHO Base    & 0.167 & 0.147 & 0  & 3 & $-$3 \\
WB Pop Base & 0.187 & 0.207 & 3  & 0 & $+$3 \\
WB CM Base  & 0.473 & 0.487 & 5  & 3 & $+$2 \\
WB Mat Base & 0.787 & 0.840 & 8  & 0 & $+$8 \\
\bottomrule
\end{tabular}%
}
\end{table}

The maximum naming bias is $+$8/150$=$5.3\% (WB Mat), and the
direction \textit{favors} the Baseline---value-first finds more
Baseline facts, not fewer. For SGE graphs, both protocols yield
identical FC (all rescued=lost=0) because SGE's structured binding
ensures that entity, value, and year co-occur in the same 2-hop
neighborhood regardless of lookup direction.

\textbf{Summary.} All three controls converge: (1)~fuzzy matching
does not shrink the gap; (2)~Baseline failures are caused by
missing values, not missing entity access; (3)~reversing the lookup
direction yields equal or higher Baseline FC. The downstream
accuracy difference reflects genuine binding fidelity, not naming
bias.

\subsection{Evaluation Blindness: Per-System Survey}\label{app:eval-blindness-survey}

Table~\ref{tab:eval-survey} summarizes how each of the 6 major
GraphRAG systems evaluates construction quality. All rely exclusively
on end-to-end QA accuracy; none measures graph-level fidelity
(e.g., fact coverage, triple F1, or edge correctness).

\begin{table}[H]
\centering
\caption{Evaluation methods used by major GraphRAG systems. None
includes a graph-fidelity metric; all evaluate via downstream QA
only.}
\label{tab:eval-survey}
\resizebox{\columnwidth}{!}{%
\begin{tabular}{llcl}
\toprule
\textbf{System} & \textbf{Evaluation Metric(s)} & \textbf{Graph Fidelity?} & \textbf{Eval Datasets} \\
\midrule
MS GraphRAG \citep{Edge2024} & Comprehensiveness, Diversity, Win rate (LLM judge) & \xmark & Podcast, News \\
LightRAG \citep{Guo2024} & Accuracy via LLM judge (4 modes) & \xmark & Legal, Agriculture, CS, Mix \\
HippoRAG \citep{Gutierrez2024} & QA Accuracy, F1 (MuSiQue, 2Wiki, HotpotQA) & \xmark & Multi-hop QA \\
nano-GraphRAG & Same as LightRAG (fork) & \xmark & (no published evaluation) \\
RAG-Anything \citep{Guo2025b} & QA Accuracy via LLM judge & \xmark & Multimodal docs \\
AutoSchemaKG \citep{Bai2025} & QA Accuracy, Schema quality (LLM judge) & \xmark & Web corpora \\
\bottomrule
\end{tabular}%
}
\end{table}

This uniform reliance on E2E QA means that upstream construction
regressions---such as the catastrophic mismatch documented in
\S\ref{sec:mismatch}---can be missed by downstream-only evaluation. Our
CSVFidelity-Bench addresses this gap by providing deterministically
verifiable ground truth at the graph construction layer.

\end{document}